\documentclass[11pt]{article}

\usepackage[final]{acl}

\usepackage{times}
\usepackage{latexsym}

\usepackage[T1]{fontenc}
\usepackage[utf8]{inputenc}

\usepackage{microtype}

\usepackage{inconsolata}

\usepackage{hyperref}
\usepackage{url}
\usepackage{graphicx}
\usepackage{wrapfig}
\usepackage{booktabs}
\usepackage{multirow}
\usepackage[many]{tcolorbox}
\usepackage{listings}
\usepackage{subcaption}
\usepackage{arydshln} 
\usepackage{pifont}
\usepackage{textcomp}
\usepackage{amsthm}
\usepackage{amsmath}
\usepackage{amssymb}
\usepackage{amsfonts}
\newtheorem*{remark*}{Remark}

\lstdefinestyle{pythonstyle}{
    language=Python,
    basicstyle=\ttfamily\small,   % 设置代码字体
    keywordstyle=\color{blue},     % 关键字颜色
    commentstyle=\color{gray},     % 注释颜色
    stringstyle=\color{red},       % 字符串颜色
    numbers=none,                  % 去掉行号
    backgroundcolor=\color{gray!10}, % 设置背景颜色
    frame=single,                  % 设置代码框线
    rulecolor=\color{black},       % 框线颜色
    showstringspaces=false,        % 不显示字符串中的空格
    breaklines=true,               % 自动换行
}

\newtcolorbox[auto counter]{promptbox*}[2][]{%
  colframe=teal!80!black, colback=teal!5, coltitle=white,
  sharp corners, boxrule=0.8mm, width=\textwidth,
  halign=left, valign=center, left=0.8mm, right=0.8mm,
  title=Prompt~\thetcbcounter:~#2,   % 标题里显示编号
  label={box:prompt\thetcbcounter},  % 自动生成唯一 label（给 \ref 用）
  #1
}

\newcommand{\nickname}{AdaR}

\title{AdaR: A Framework for Equipping LLMs with Adaptive Reasoning}

\author {
    Zhejian Lai\textsuperscript{\rm 1 \rm2}\footnotemark[1] \quad
    Xiang Geng\textsuperscript{\rm 1}\footnotemark[1] \quad 
    {\bf Zhijun Wang}\textsuperscript{\rm 1} \quad
    {\bf Bai Yang}\textsuperscript{\rm 2} \quad
    {\bf Jiahuan Li}\textsuperscript{\rm 2} \quad \\
    {\bf Rongxiang Weng}\textsuperscript{\rm 2} \quad
    {\bf Jingang Wang}\textsuperscript{\rm 2} \quad
    {\bf Xuezhi Cao}\textsuperscript{\rm 2} \quad
    {\bf Xunliang Cai}\textsuperscript{\rm 2} \quad
    {\bf Shujian Huang}\textsuperscript{\rm 1} \footnotemark[2] \\
    \textsuperscript{\rm 1} National Key Laboratory for Novel Software Technology, Nanjing University, Nanjing, China\\
    \textsuperscript{\rm 2} Meituan, Beijing, China\\
    \texttt{\{laizj, gx, wangzj\}@smail.nju.edu.cn,\{huangsj\}@nju.edu.cn} \\
    \texttt{\{baiyang28, lijiahuan04, wengrongxiang\}@meituan.com} \\
    \texttt{\{wangjingang, caoxuezhi, caixunliang\}@meituan.com}
}

\begin{document}
\maketitle

\renewcommand{\thefootnote}{\fnsymbol{footnote}}
\footnotetext[1]{Equal contribution, in random order.}
\footnotetext[2]{Corresponding author.}
\renewcommand{\thefootnote}{\arabic{footnote}}
\begingroup
\renewcommand{\thefootnote}{}
\footnotetext{Accepted to the Findings of the 2026 Conference on
Empirical Methods in Natural Language Processing (EMNLP 2026).}
\addtocounter{footnote}{-1}
\endgroup

\begin{abstract}
Mathematical reasoning is a primary indicator of large language models (LLMs) intelligence.
However, existing LLMs exhibit failures of robustness and generalization.
This paper attributes these deficiencies to spurious reasoning, wherein generated reasoning traces are driven by superficial correlations, leading models to blindly reproduce memorized patterns from the training data.
% generated reasoning traces bear negligible causal correlation to answers.
% producing answers from superficial features.
To address this challenge, we propose the {\nickname} framework to equip LLMs with adaptive reasoning, wherein models establish correct correlations between query template and reasoning process.
{\nickname} automatically synthesizes logically equivalent queries by varying variable values, and trains models with Reinforcement Learning with Verifiable Rewards (RLVR) to penalize spurious logic while encouraging adaptive logic.
To ensure data quality, we extract the problem-solving logic from the original query and generate the corresponding answer by code execution and then apply sanity check.
Experimental results demonstrate that {\nickname} achieves substantial improvement in mathematical reasoning while maintaining high data efficiency.
Furthermore, even for advanced LLMs, there still exists robustness and generalization deficiencies, which our work effectively mitigates.
% Analysis indicates that perturbing variable values is more effective than paraphrasing to improve model's performance.
% Analysis indicates that data synthesis and RLVR function in a coordinated manner to equip LLMs with adaptive reasoning.
% Subsequent analyses derive key design insights into the effect of critical factors and the applicability to instruct LLMs.
Our project is available at \url{https://github.com/NJUNLP/AdaR}.
\end{abstract}

\section{Introduction}

Large Language Models (LLMs) have demonstrated strong performance across various reasoning tasks~\citep{cot}. 
Among these, mathematical reasoning serves as a crucial cognitive skill that supports problem-solving across tasks~\citep{math_survey}. 
Beyond early direct inference attempts~\citep{liu2021pretrainpromptpredictsystematic, brown2020language}, Chain-of-Thought (CoT) reasoning has been recognized as an effective approach to enhance mathematical reasoning~\citep{wei2022chain}, as it breaks down complex problems into manageable steps~\citep{chu2023navigate}.

% gx：
% SFT不要单独说了，直接说LLM有spurious COT的情况，spurious COT获得不了这些好处，然后按照现在的逻辑讲
% 然后我们argue adaptive的COT能获得
% 然后接上一句说目前的RL方法也不能解决这个问题，然后按照现在的逻辑讲

\begin{figure}[tbp]
    \centering    \includegraphics[width=\columnwidth]{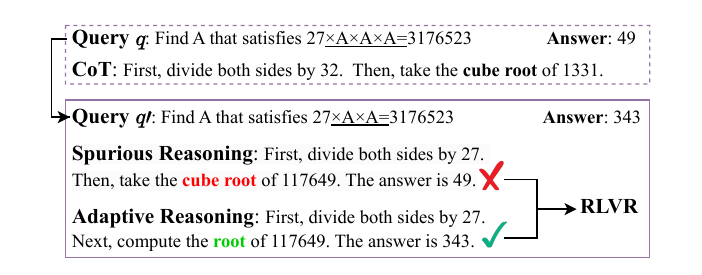}
    \caption{
    The examples of spurious and adaptive reasoning.
    }
    \label{fig:example}
    \vspace{-10pt}
\end{figure}

However, existing mathematical LLMs still exhibit failures at two levels: (i) robustness on in-domain tasks
% , the performance of models exhibit a noticeable performance drop when confronted with numerical variations of problems
~\citep{gsm_sym};
and (ii) generalization on out-of-domain tasks
% , models fine-tuned on math word problems perform poorly on theorem-based reasoning tasks
~\citep{jahin2025evaluating}.
We attribute these deficiencies to \textbf{spurious reasoning}, which occurs when models establish incorrect correlations during training.
As shown in Figure~\ref{fig:example}, when conditions are changed, the model still blindly outputs the original reasoning token (e.g., ``cube''), demonstrating that model's reasoning generation depends on memorized variable values.
% , rather than structural query template $T$  which truly governs the problem-solving logic $L$.
% where LLMs derive the answer $y$ from superficial patterns rather than the underlying problem-solving logic $L$, yielding CoTs that are semantically similar but logically inequivalent to correct ones. As shown in Figure~\ref{fig:example}, the model solves query in training set correctly, but after only numerical values are changed, it produces a similar CoT that omits a key reasoning token, e.g., ``cube'', thereby altering the problem-solving logic and producing an incorrect answer. Thus, even with an unchanged query template, models relying on spurious reasoning fail to adapt to changes in the variable set $x$.
% (new)和generalization的关系
Due to the spurious correlations, the whole spurious reasoning trace cannot be explained from atom of thoughts~\citep{teng2025atom}.
Therefore, it's non-compositional along causal relations which causes  poor generalization of the models.

% 根据代数思想，一个数学问题（甚至是通用的推理问题），由问题描述（Template）和变量集组成，（T(x)）
% 如果期望模型的推理路径随着变量集的改变而适应性的调整，就需要将解答逻辑建模为一个关于变量集的函数。(L(x))
% 最终可以得到正确的y。

\begin{figure*}[tbp]
    \centering    \includegraphics[width=\linewidth]{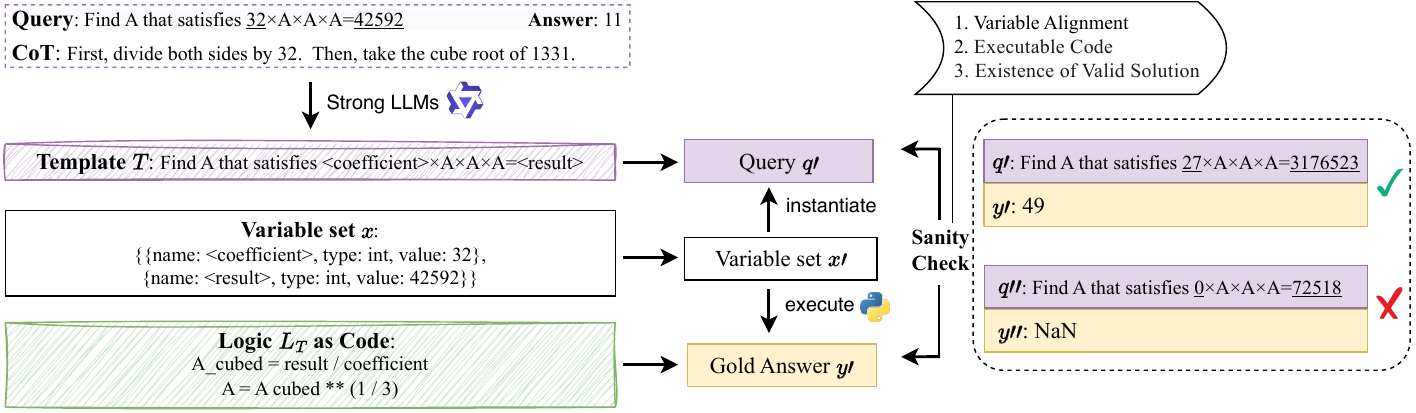}
    \caption{
    % The overview of {\nickname}. 
    The pipeline of how we get the query-answer pairs by controllably perturbing variable values while preserving problem-solving logic and sanity.
    }
    \label{fig:process}
    \vspace{-10pt}
\end{figure*}

% We argue that an genuine reasoning process, i.e. \textbf{adaptive reasoning}, should be able to understand and solve $q$ by $L$, while be adaptive to changeable values in $x$.

To overcome the mentioned weaknesses, the model is expected to establish a robust correlation to the query template $T$ where specific variables $x$ are disentangled from the query $q$ — an ability we define as \textbf{adaptive reasoning}.
% We introduce this structure-to-token generative process as \textbf{adaptive reasoning}.

% To overcome the mentioned weaknesses, the model is expected to understand the logic $L$ as a function $L(.)$, by correctly understanding the query $q$ and disentangling $x$ from $q$, which is an ability that reflects algebraic reasoning~\citep{kieran2004algebraic}. 
% Applying the disentangled $x$ to $L(.)$ may reveal the correct answer $y$.
% % By applying the disentangled $x$ into $L(x)$, the model acquire the final answer $y$.
% We introduce this process as \textbf{adaptive reasoning}.
% The query template is denoted as $T$, with $x$ removed from $q$. %So each $T$ implies an $L$.

To compel the model to establish the correct correlation to templates, it must be exposed to diverse data that systematically varies the values in variable set $x$ as well as the textual expression of the template $T$, while keeping the corresponding problem-solving logic $L_T$ fixed.
While previous works struggle to systematically vary both $x$ and $T$ while strictly maintaining high-quality, logic-consistent reasoning traces.
% safely and jointly scale $x$ and $T$  dimensions. 
For instance, \citet{yu2023metamath} predominantly concentrated on perturbing $T$ (e.g., paraphrasing) while keeping $x$ fixed. %while largely neglecting systematic perturbations of $x$. 
However, training with varing $T$ with fixed $x$ may increase the risk of spurious reasoning overfitting to $x$\footnote{For a formal theoretical proof, please refer to Appendix \ref{apx:theory}: Spurious Correlations \& The Limit of Varying Templates.}.
% (A theoretical analysis is shown in Appendix~\ref{apx:theory})
In contrast, \citet{lu2024mathgenie} attempted to prompt the LLM to modify $x$; however, they leaves $x$ uncontrolled, and may inadvertently alter $L_T$ due to hallucination~\cite{huang2025survey}, making it difficult to ensure the correctness of generated gold answer $y$.

In order to equip LLMs with \textbf{Ada}ptive \textbf{R}easoning, we propose the framework \textbf{\nickname{}},  which synthesizes data along both $T$ and $x$ dimensions and performs reinforcement learning with them. 
Crucially, \nickname{} introduces an automatic process for high-precision, controllable numerical perturbations, while ensuring verified $y$.
% {\nickname} synthesizes diverse data by keeping the problem-solving logic unchanged and perturbing the values in the variable set.
%Two challenges must be solved when performing perturbations: preserving sanity while perturbing, and obtaining the gold answer without human annotations.
To achieve this end, we decompose the complex perturbation task into the following manageable, verifiable sub-tasks.
As shown in Figure~\ref{fig:process}, we first prompt a strong LLM to generate $T$, $L_T$ (as code, e.g. a Python program), and $x$.
Subsequently, we \textit{controllably perturb} values in $x$ to predefined magnitudes and types.
The perturbed variable set $x'$ is then used to instantiate $T$ to generate queries $q'$; while $x'$ is also provided as input to $L_T$, which is executed to produce $y'$.
Furthermore, we introduce \textit{sanity check} to filter invalid instances.

% RL怎么通过提供正负反馈进行的对比，扰动数据在其中的重要性 (信号建立在扰动数据之上)
% 训练方法怎么利用合成数据
{\nickname} then trains the model by Reinforcement Learning with Verifiable Rewards (RLVR)~\citep{shao2024deepseekmath} to equip LLMs with adaptive reasoning. %so that the model can compare the feedback obtained from solving queries of these data.
In the single-data setting, outcome rewards are unreliable, since spurious reasoning may still lead to correct answers through superfical patterns. However, when inferring on perturbed queries from {\nickname},  spurious reasoning is unlikely to remain valid for all these queries, so outcome correctness provides a more reliable reward, penalizing spurious reasoning and pushing the model to explore the adaptive problem-solving logic.

% Notably, unlike in the single data situation, where outcome reward may inadvertently strengthen spurious reasoning, %{\nickname} can synthesize data conditioned on that data.
% the correctness of outcomes on these synthetic data provides a reliable signal for inferring where their responses derive.
% %In detail, when solving queries of these data, in contrast to adaptive reasoning, responses that persist in relying on spurious reasoning are more likely to produce incorrect answers on these synthetic data and are consequently penalized in RLVR thereby encouraging the model to explore the problem-solving logic (as shown in subfigure III of Figure~\ref{fig:process}).
% In detail, responses that rely on spurious reasoning are more likely to produce incorrect answers on the perturbed synthetic data and are consequently penalized in RLVR, thereby pushing the model to explore the adaptive problem-solving logic (Figure~\ref{fig:example}).

Extensive experiments across in-domain robust tasks and out-of-domain tasks demonstrate that {\nickname} achieves great gains (+8.50 points on average), with only 9K synthetic data on both math-specialized and general base models, thereby demonstrating that our approach enhances the model's robustness and generalization.
Further analysis indicates that:
\begin{itemize}
    \item Advanced LLMs still exhibit pronounced robustness and generalization deficiencies, which our work effectively mitigates.
    \item Perturbing the variable values is more effective than perturbing the query template for improving model's performance.
    \item Evidence from attention towards templates and improved algebraic thinking  demonstrates that {\nickname} equips LLMs with adaptive reasoning
    %\item Contrast between responses generated by both spurious and adaptive reasoning improves performance.
\end{itemize}

\vspace{-5pt}
\section{Method}
\vspace{-5pt}
% insights and overview 
% related work里面要说 相比随机扰动后让LLM得到标签，我们的方法标签更准。相比直接让模型生成答案
% 保证利用了问题变体中模型使用捷径进行推理的负样本，以使得模型掌握避免学习捷径，掌握推理逻辑。

In this section, we first present methods for synthesizing data that ensure controllability and sanity throughout the generation process. 
Following this, we introduce our training strategy, which is employed to more effectively integrate with synthetic data, preventing the model from learning spurious reasoning, thereby facilitating adaptive reasoning.

\subsection{Data Synthesis with Executable Code and Verifiable Answers}

A straightforward way to perturb values in queries and obtain correct answers without human annotation is to directly prompt a LLM~\citep{wang2025comprehensivesurveydataaugmentation}.
However, our preliminary experiments show that such approach provides limited control over neither the magnitudes nor types of perturbations, and cannot guarantee answer correctness.
To address these limitations, {\nickname} decomposes data synthesis into a sequence of manageable, verifiable sub-tasks. Specifically, the LLM is only tasked with extracting necessary metadata within the data, including query template and variable set, and translating the given CoT into code; controllable perturbations are then applied externally, and sanity check are performed to ensure the correctness of synthesized query-answer pairs.
We detail each sub-task below.

\paragraph{Convert Logic in Text to Logic in Code.}
Executable problem-solving code can serve as a problem-solving logic provided its correctness is guaranteed.
Crucially, such code generalizes effectively: by substituting the input variable set, it can solve any perturbed query of the original query, for its output can serve as reliable gold answer.
When CoT and gold answer are provided, problem-solving code generation becomes a straightforward process of translating the textual logic (i.e. CoT) into executable code, thereby reducing the model’s reasoning burden and enabling direct verification of correctness against the gold answer.
Building on these insights, we provide a query $q$, the corresponding CoT $z$ and gold answer $y$ to an open-source LLM to synthesize the logic $L_T$ as code for producing gold answer y' of subsequent perturbed query $q'$.
Because code generation requires abstracting the concrete numerical values in $q$ into a mapping to input variables, we further instruct the model to produce query template $T$ as a byproduct by applying this mapping into the original query $q$.
Appendix~\ref{apx:prompt} details the employed prompt .

% , we prompt LLMs to automatically generate both the executable code and, as a by-product, extract the mapping between the code variables and the numerical descriptions in the given problem. 
% With this mapping, we substitute the concrete numerical values in the original problem with placeholders, denoted as \textit{\textless variable\textgreater}, thereby generating a problem template.

% \begin{tcolorbox}[
%   colframe=teal!80!black, % 框的颜色
%   colback=teal!5,          % 背景颜色
%   coltitle=white,          % 标题的颜色
%   title=Example 3.1: Code and Problem Template, % 标题内容
%   sharp corners,           % 更尖锐的角
%   boxrule=0.8mm,           % 框线的粗细
%   width=\textwidth,     % 框的宽度
%   halign=left,           % 内容居中
%   valign=center,            % 内容垂直居中
%   left=0.8mm, right=0.8mm,
% ]
% \label{exp:template}
% \textbf{Problem:}
% Find A that satisfies 32×A×A×A=42592

% \textbf{Code:}
% \begin{lstlisting}[style=pythonstyle]
% # Variable definitions
% coefficient = 32
% result = 42592

% # Calculation
% A_cubed = result / coefficient
% A = A_cubed ** (1/3)

% # Output
% print(A)
% \end{lstlisting}

% \textbf{Template:}
% Find A that satisfies \textit{\textless coefficient\textgreater}×A×A×A=\textit{\textless result\textgreater}
% \end{tcolorbox}

\paragraph{Controllable Perturbation.}
By comparing $q$ with $T$, we construct a variable set $x$ that records each variable's name, numeric type and its value as it appears in the original query.
Since the values are explicitly extracted, perturbations can be applied directly at a predefined magnitude while ensuring numerical validity, thereby avoiding the unpredictability of perturbations performed through an LLM.
More specifically, we then apply independent perturbations by sampling each variable's value within a range of $\pm\alpha\%$ of its original value.
The parameter $\alpha$ allows the control of the perturbation magnitude.
To ensure numerical validity, both the numeric type and the sign of every variable must remain unchanged before and after perturbation.
Formally, for the \(i\)-th variable with original value \(x_0^{i}\), we draw:
\begin{equation}
\label{eq:perturb}
\begin{aligned}
x^i = x^i_0 \times (1 + \Delta_i), \Delta_i \sim \text{Uniform}(-\alpha\%, \alpha\%)\\
\text{s.t.}\ \text{type}(x^i) = \text{type}(x^i_0),\ \text{sign}(x^i) = \text{sign}(x^i_0)
\end{aligned}
\end{equation}

\paragraph{Sanity Check.}
\label{sec:sanity}
After applying perturbations, we instantiate $T$ with $x'$ to obtain $q'$, and execute $L_T$ as code with the same $x'$ as input to generate $y'$.
Although providing $z$ and $y$ has reduced task difficulty, $T$ and $L_T$ as code generated by LLMs remain uncertain and may contain errors. 
Furthermore, a meaningful query imposes inter-variable constraints on its variables; independent perturbations can violate these constraints and thereby introduce errors.
To ensure that the perturbed data remain well-posed, we conduct a \textit{sanity check} along the following aspects:
\begin{itemize}
    \item \textbf{Variable Alignment (VA).} We compare variables referenced in $T$ with those used in each $L_T$ as code. 
    Any mismatch indicates potential errors (e.g. hallucination).
    \item \textbf{Executable Code (EC).} Since $L_T$ as code is used to derive the $y$, executability is a critical requirement: (i) the code runs without runtime errors; and (ii) providing the original variable set $x_0$ as input reproduces the gold answer $y_0$.
    \item \textbf{Existence of Valid Query (EVQ).} 
    As independent perturbations may violate implicit inter-variable constraints, they can produce invalid queries (e.g., selecting $20$ items from a set of $10$).
    To provide evidence that a valid query exists, we cross-validate the code-executed answer with the answer produced by a mathematical LLM.
    This validation is based on the following rationale. LLMs are prone to hallucinate under out-of-distribution inputs (i.e. invalid queries)~\citep{huang2025survey}. Given the large solution space of mathematical queries, exact agreement between the two answers is unlikely unless the query is valid and both code and the LLM follow the correct logic. The latter condition is implemented because EC ensures that $L_T$ as code is correctly translated from provided CoT, meanwhile we provide  $L_T$ as contextual guidance to help LLM align with correct logic when inferring. We therefore use answer agreement as a evidence of the existence of a valid query.

    % As noted by \citep{huang2025survey}, LLMs are prone to hallucinate under out-of-distribution settings. 
    % To provide evidence that a valid solution exists, we perform cross-validation by comparing the answer generated by the code with the output of a mathematical LLM under perturbed query input. 
    % To handle this, we supply the corresponding $L_T$ as code in the context as a hint, enabling the model to optionally ground its reasoning on $L_T$.
\end{itemize}
If a perturbed instance fail to pass the \textit{sanity check}, we reattempt \textit{controllable perturbation}. 
If the number of attempts exceeds $\tau$ times, we conclude this instance likely involves complex inter-variable constraints and then discard it from synthesis.

\paragraph{Queries Paraphrasing.}
We adopt paraphrasing~\citep{yu2023metamath} as a data augmentation strategy to increase the diversity of $T$. 
This approach complements the \textit{controllable perturbation} introduced earlier to increase the diversity of $x$.
The impact of these scalable dimensions on model performance is further explored in Section~\ref{sec:ablation_study} and Section~\ref{sec:diversity}.

% 有点突兀有点怪
\begin{remark*}
Our synthetic data contains no CoT, it consists solely of the query and the corresponding gold answer.
% 外部提供的CoT与target LLM的reasoning无关,进而无法帮助模型master adaptive reasoning，相反只有采样出来的CoT才能帮助模型
As will be described in the next subsection, CoTs generated through either spurious reasoning or adaptive reasoning can be sampled from the target LLM.
\end{remark*}

\subsection{Training Strategy}
The training objective of Supervised Fine-Tuning (SFT) and its variant Rejection sampling Fine-Tuning (RFT)~\cite{yuan2023scaling} (detailed in Appendix~\ref{apx:obj}) makes models prone to memorizing provided CoTs rather than developing adaptive reasoning~\citep{chu2025sft}, thereby causing them to converge to local optimum~\cite{kang2025quagmires}.
RLVR has recently been widely adopted to enhance generalization~\citep{guo2025deepseek}.
However, its outcome reward generated by spurious reasoning is indistinguishable from that generated by adaptive reasoning, which may inadvertently reinforce spurious reasoning.

To address this, we combine RLVR with our synthetic data. 
RLVR's reward-driven enables the model to learn from comparison among rewards obtained by solving perturbed queries using different reasoning process. 
More specifically, when given the original query $q$, it's hard to determine where the gold answer $y$ is derived.
However, when the model is evaluated on perturbed query $q'$, reliance on CoT from spurious reasoning is more likely to yield incorrect outcome, whereas reliance on CoT from adaptive reasoning $L_T(.)$ is more likely to yield correct outcome.
All perturbed queries are placed into the same batch.
Then, each reasoning process is more likely to receive appropriate feedback, thereby promoting adaptive reasoning.
% Therefore, we combine RLVR with perturbed queries $q_i^j$, under which reliance on spurious reasoning is likely to yield incorrect outcomes $\hat{y}_i^j$.
% Then, such inferences are penalized, thereby promoting the adaptive reasoning, as shown in subfigure III of Figure~\ref{fig:process}.

\section{Experiment}
\subsection{Experimental Setup}

\paragraph{Data synthesis.} We use the Qwen2.5-72B-Instruct~\citep{yang2024qwen2} as the open-source LLM to generate the query templates and the problem-solving codes.
We select $9K$ instances from ORCA-MATH~\citep{orca-math} as seed data for data synthesis.
Using these seed data, {\nickname} synthesizes instances with a predefined magnitude $\alpha=500$ and a maximum of $\tau=50$ attempts.
For each data in seed data, we select one corresponding synthetic data to construct the ``ORCA-{\nickname}-train'', which contains $9K$ instances. 
We select another $2.5K$ instances to form ``ORCA-{\nickname}-test'', ensuring no overlap with ``ORCA-{\nickname}-train''. 
% ORCA-{\nickname}-train
% ORCA-{\nickname}-test
% variable是unseen
The details are shown in Appendix~\ref{apx:data_syn}.

\paragraph{Models.}
% 选不同math和general的models突出方法的适应性
To prove the adaptability of our framework, we conduct experiments on two categories of base models: math specialized base LLM, specifically Qwen2.5-Math-7B~\citep{yang2024qwen2} and DeepSeekMath-7B~\citep{shao2024deepseekmath}, and 8B general base LLM, specifically LLaMA3-8B~\citep{grattafiori2024llama}. 
In Section~\ref{sec:instruct}, we further explore the performance of {\nickname} on Instruct Model.

% 加个Analysis的段落，比较Instruct vs Instruct + AdaR，理论和现象是相吻合的

\begin{table*}[t]
\centering
\resizebox{\linewidth}{!}{
\begin{tabular}{c c c c c c c c c c c}
\toprule
\multirow{2}{*}{Method} & \multicolumn{5}{c}{In-Domain} & \multicolumn{4}{c}{Out-of-Domain} & \multirow{2}{*}{AVG} \\
\cmidrule(lr){2-6} \cmidrule(lr){7-10} & GSM8K & ORCA-{\nickname} & main & p1 & p2 & MATH & College & Theorem & AIME & \\
\midrule
\multicolumn{11}{c}{\textbf{Qwen2.5-MATH (7B MATH-Specialized Base Model)}} \\
\midrule
% Instruct & {92.57} & {81.44} & {89.60} & {81.86} & {70.68} & {79.86} & {51.70} & {44.25} & {13.75} & 67.30 \\
Initial SFT & 81.43 & 77.12 & 74.26 & 64.48 & 52.44 & 42.84 & 28.55 & 14.88 & \ \ 4.27 & 48.92\\
\hdashline
%Standard-SFT & {81.12} & {78.56} & {74.86} & {65.14} & {52.36} & {43.00} & {29.99} & {14.38} & {\ \ 6.67} & {49.56} \\
Standard-RLVR & {86.96} & {81.80} & {81.34} & {74.32} & {64.68} & {61.74} & {29.48} & {20.38} & {10.94} & {56.85} \\
MetaMATH &{84.61} & {82.44} & {79.78} & {71.24} & {63.28} & {66.92} & {39.25} & {26.13} & {\ \ 9.38} & {58.11} \\
MathGenie &{87.64} & {84.96} & {80.40} & {72.60} & {63.56} & {55.94} & {21.11} & {20.05} & {10.31} & {55.17} \\
{\nickname} & \textbf{91.81} & \textbf{86.08} & \textbf{89.72} & \textbf{82.74} & \textbf{73.24} & \textbf{75.90} & \textbf{48.62} & \textbf{37.13} & \textbf{14.27} & \textbf{66.61} \\
\midrule
\multicolumn{11}{c}{\textbf{DeepSeekMath (7B MATH-Specialized Base Model)}} \\
\midrule
% Instruct & {79.30} & {59.28} & {62.16} & {57.50} & {29.28} & {32.46} & {29.28} & {17.75} & {0.21} & {40.80} \\
%Standard-SFT & {72.63} & {69.32} & {72.32} & {50.22} & {25.64} & {29.26} & {23.85} & {11.75} & {0.00} & {39.44} \\
Initial SFT & 70.28 & 65.48 & 61.70 & 50.26 & 29.36 & 28.44 & 22.32 & 11.50 & 0.21 & 37.73 \\
\hdashline
Standard-RLVR & {81.43} & {72.56} & {74.76} & {64.34} & {40.00} & {39.26} & {21.61} & {20.38} & {0.73} & {46.12} \\
MetaMATH &{80.89} & {70.28} & {75.14} & {64.54} & {39.56} & {38.02} & {19.13} & {22.63} & {0.21} & {45.60} \\
MathGenie &{80.13} & {73.44} & {74.18} & {64.02} & {40.04} & {39.16} & {22.39} & {23.75} & \textbf{3.33} & {46.72} \\
{\nickname} & \textbf{81.43} & \textbf{75.44} & \textbf{76.56} & \textbf{64.94} & \textbf{42.00} & \textbf{41.18} & \textbf{24.23} & \textbf{25.25} & \textbf{3.33} & \textbf{48.26} \\
\midrule
\multicolumn{11}{c}{\textbf{Llama3 (8B General Base Model)}} \\
\midrule
% Instruct & {79.38} & {60.28} & {78.02} & {69.90} & {50.36} & {47.72} & {30.73} & {20.75} & {2.50} & {48.85}  \\
%Standard-SFT & {69.52} & {54.58} & {64.10} & {47.80} & {27.52} & {18.16} & {\ \ 9.26} & {\ \ 9.50} & {0.00} & {33.38} \\
Initial SFT & 67.17 & 57.84 & 59.98 & 47.30 & 24.92 & 18.20 & \ \ 9.33 & \ \ 8.13 & 0.00 & 32.54\\
\hdashline
Standard-RLVR & {73.09} & {55.68} & {67.06} & {51.58} & {25.16} & {19.02} & {\ \ 9.01} & {\ \ 9.00} & {0.00} & {34.40} \\
MetaMATH & {74.52} & {59.96} & {70.74} & {56.58} & {26.72} & {19.68} & {\ \ 9.04} & {\ \ 9.38} & {0.00} & {36.29} \\
MathGenie & {72.47} & {58.90} & {67.26} & {53.72} & {25.38} & {18.30} & {\ \ 8.92} & {\ \ 9.00} & {0.00} & {34.88} \\
{\nickname} & \textbf{77.77} & \textbf{61.52} & \textbf{73.96} & \textbf{56.90} & \textbf{30.28} & \textbf{20.48} & \textbf{\ \ 9.62} & \textbf{\ \ 9.63} & {0.00} & \textbf{37.80} \\
\bottomrule
\end{tabular}
}
\caption{
 %Performance comparison on mathematical benchmarks including GSM8K, ORCA-{\nickname}-test (ORCA-{\nickname}), GSM-SYM (main/p1/p2), CollegeMATH(College), OlympiadBench-Math (Olympiad), TheoremQA (Theorem), and AIME 2025 (AIME).
 Performance comparison across In-Domain and Out-of-Domain mathematical benchmarks. The datasets ``main'', ``p1'', and ``p2'' are from GSM-SYM. Best results are highlighted in bold.
}
\label{tab:main_result}
\vspace{-10pt}
\end{table*}

\paragraph{Evaluation.} For a comprehensive evaluation of mathematical reasoning, we adopt 7 benchmarks covering both in-domain and out-of-domain evaluation. 
Specifically, we evaluate in-domain mathematical competence using GSM8K~\citep{gsm8k} and evaluate in-domain robustness using ORCA-{\nickname}-test (ORCA-{\nickname}) and GSM-SYM (main/p1/p2)~\citep{gsm_sym}. Within GSM-SYM, the main, p1, and p2 subsets exhibit increasing difficulty.
For out-of-domain evaluation, we use MATH~\citep{MATH}, CollegeMath~\citep {colleage_math}, TheoremQA~\citep{theoremqa}, and American Invitational Mathematics Examination (AIME) problems for 2025 to evaluate generalization.
Results are reported as pass@1 for all datasets except AIME 2025, for which, in accordance with \citep{yu2025dapo}, we report avg@32.
Further details about the evaluation setup and benchmarks are provided in the Appendix~\ref{apx:eva}.

\paragraph{Baseline.}
We primarily compare our framework, {\nickname}, against the following baselines:
(1) 
% ``Instruct'' versions of base model trained on large-scale data, which first undergo CoT-based SFT and are subsequently optimized by RL.
The base models undergo SFT using 9K seed data—this same data is also leveraged for data synthesis. These models are referred to as the ``Initial SFT'' models.
(2) 
% Models that are instruction-tuned on an 18K subset of the ORCA dataset, denoted as the ``Standard-SFT'' setting; and
% 低优：解释一下这里9K不一样
In ``Standard-RLVR'' setting, the ``Initial SFT'' is subsequently trained with RLVR on another 9K subset of ORCA-MATH, with its templates being disjoint from the training set used for ``Initial SFT''.
(3) 
% related works里说
% Since our framework does not rely on CoTs distilled from powerful LLMs and aims to improve the generalization capabilities for enhanced reasoning, we reproduce related state-of-the-art methods for comparison.
% 低优: 说的清楚才说
% 直接说比了其他数据扰动方法
We also include other synthesis methods, such as MetaMATH~\citep{yu2023metamath}, which improves query diversity through paraphrasing and self-verification; MathGenie~\citep{lu2024mathgenie}, which generate questions by applying back-translation to paraphrased responses. 
For all RLVR training strategy, we adopt DAPO~\citep{yu2025dapo}. 
% 可删可不删的一句话
% for its faster convergence and the elimination of the need for a value model, which together reduce training time and computational resource requirements.
Additional details of the training setup are provided in the Appendix~\ref{apx:training}.

\subsection{Main Results}
The main results are summarized in Table~\ref{tab:main_result}. 
We highlight the following three observations:
\paragraph{Observation 1: Our method with a small amount of synthetic data yields substantial performance gains.} 
Using only 9K synthetic data, our method surpasses other methods across all base models.
Compared to other synthetic data methods, we improve over MetaMATH by 8.50 points and over MathGenie by 11.44 points on average.

\label{sec:main_result_2}
\paragraph{Observation 2: Our method significantly enhances model robustness and generalization.}
% seen和unseen的是什么T,x,L？扰动的是什么？
% 你在前面只说是robustness测试集，不要详细讲，在这讲

The enhancement manifests across three levels: (i) perturbed variable values in seen queries during training (ORCA-{\nickname}-test); (ii) perturbed variable values in unseen queries during training (GSM-SYM main/p1/p2, where main, p1, and p2 represent increasing levels of difficulty); (iii) out-of-domain data.
% 相比于次优的
Across all base models, we observe gains on the first two levels compared to the suboptimal MetaMATH ($+4.66$ / $+4.75$ points).
% The  main test set comprises $100$ query templates; for each template, $500$ instantiations constitute a group.
% If the model master adaptive reasoning, it implies it can effectively solve all queries in one group and form a fully correct groups, since the problem-solving logic in one group is unchanged.
% However, the spurious reasoning exhibits unstable performance, producing mixed groups that contain both correct and incorrect answers.
At a finer level of granularity, we observe that ``Initial SFT'' rarely produces correct answers across all perturbations of variable values within a specific query template.
In contrast, {\nickname} significantly improves this capability.
% we observe that AdaR makes $38.71\%$ of the answers to  template-derived perturbed queries from containing errors to being entirely correct.
% 更细粒度的, 在GSM-SYM中， we find that our method 让 模型在回答 $38.71\%$ 模版得到的perturbed queries 从 包含有错误 变成全对.
Besides, on the third level we obtain larger improvements:
when using Qwen2.5-MATH-7B as the base model, the average gain reaches +8.56 points.
These results indicate that the model generates CoTs from adaptive reasoning rather than spurious reasoning, which better supports robustness and generalization.

% 这句话没有证据，但比较直觉

% \begin{table*}[t]
% \centering
% \caption{
% Ablation Study.
% }
% \resizebox{0.8\linewidth}{!}{
% \begin{tabular}{lcccccccccc}
% \toprule
% \multirow{2}{*}{Method} & 
% \multirow{2}{*}{Strategy} &
% \multicolumn{3}{c}{Sanity Check} & \multirow{2}{*}{Paraphrase} & \multirow{2}{*}{In-Domain} & \multicolumn{4}{c}{Out-of-Domain} \\
% \cmidrule{3-5} \cmidrule{8-11}
% & & VA & EC & EVS & & &  MATH & College & Theorem & AIME \\
% \midrule
% Initial SFT & - & - & - & - & - & 69.95 & 42.84 & 28.55 & 14.88 & \ \ 4.27 \\
% \hdashline
% \multirow{6}{*}{{\nickname}} & RFT & \ding{51} & \ding{51} & \ding{51} & \ding{51} & 72.91 & 45.20 & 30.41 & 16.13 & \ \ 3.54  \\
% & RLVR & \ding{55} & \ding{51} & \ding{51} & \ding{51} & 82.32 & 71.58 & 44.14 & 32.50 & \ \ 7.92 \\
% & RLVR & \ding{51} & \ding{55} & \ding{51} & \ding{51} & - & - & - & - & - \\
% & RLVR &\ding{51} & \ding{51} & \ding{55} & \ding{51} & 83.57 & 72.94 & 46.56 & 33.25 & \ \ 7.92 \\
% & RLVR &\ding{51} & \ding{51} & \ding{51} & \ding{55} & 84.43 & 75.28 & 48.22 & 34.88 & \ \ 6.88 \\
% & RLVR & \ding{51} & \ding{51} & \ding{51} & \ding{51} &  \textbf{84.72} & \textbf{75.90} & \textbf{48.62} & \textbf{37.13} & \textbf{14.27} \\
% \bottomrule
% \end{tabular}}
% \label{tab:ablation_study}
% \vspace{-10pt}
% \end{table*}

\begin{table*}[t]
\centering
\resizebox{0.9\linewidth}{!}{
\begin{tabular}{lccccccccccc}
\toprule
\multirow{2}{*}{Method} & 
\multirow{2}{*}{Strategy} &
\multicolumn{3}{c}{Sanity Check} &
\multirow{2}{*}{Paraphrase} &
\multicolumn{4}{c}{In-Domain} &
\multirow{2}{*}{Out-of-Domain} \\
\cmidrule{3-5} \cmidrule{7-10}
& & VA & EC & EVQ & & ORCA-\nickname{} & main & p1 & p2 & \\
\midrule
Initial SFT & - & - & - & - & - & 77.12 & 74.26 & 64.48 & 52.44 & 22.63 \\
\hdashline
\multirow{6}{*}{\nickname{}} 
& RFT  & \ding{51} & \ding{51} & \ding{51} & \ding{51} & 82.32 & 76.54 & 67.66 & 55.48 & 23.82 \\
& RLVR & \ding{55} & \ding{51} & \ding{51} & \ding{51} & 85.36 & 87.98 & 80.34 & 69.12 & 39.04 \\
& RLVR & \ding{51} & \ding{55} & \ding{51} & \ding{51} & - & - & - & - & - \\
& RLVR & \ding{51} & \ding{51} & \ding{55} & \ding{51} & 85.48 & 89.06 & 81.90 & 71.88 & 40.17 \\
& RLVR & \ding{51} & \ding{51} & \ding{51} & \ding{55} & \textbf{87.56} & 89.63 & 80.96 & 71.88 & 41.31 \\
& RLVR & \ding{51} & \ding{51} & \ding{51} & \ding{51} & 86.08 & \textbf{89.72} & \textbf{82.74} & \textbf{73.24} & \textbf{43.98} \\
\bottomrule
\end{tabular}}
\caption{
Ablation Study. Best results are highlighted in bold.
}
\label{tab:ablation_study}
\end{table*}

\paragraph{Observation 3: The effectiveness of our method correlates with the mathematical reasoning ability of the base model.}
The number of additional rollouts required to obtain both correct and incorrect answers follows the ordering
Qwen2.5-MATH(458.00) < DeepSeekMath(563.20) < LLama3(942.83),
which is inversely correlated with their respective performance gains (+17.69, +10.53, and +5.26 points).
% 需要具备采样出正负样本的基础能力作为对比，才能学会adaptive reasoning
It implies possessing sufficient mathematics-related knowledge is a necessary condition for sampling both positive and negative responses, thereby facilitating the learning of adaptive reasoning.
% 这里不好提logic的多样性，因为logic的多样性属于我们合成数据层面考虑的事情，我这里想要强调的是用RL的base model得有足够的数学能力
% 跳步骤了
% Besides, since these mathematical models have already been pre-trained on large amount real or synthetic mathematical data~\citep{yang2024qwen2,shao2024deepseekmath}, 
The result also suggests that {\nickname} is complementary to the mathematical pre-training with a large amount of real or synthetic mathematical data.
% as it further strengthens more powerful base models.

\vspace{-5pt}
\subsection{Ablation study}
\label{sec:ablation_study}
% 缺实验 三个constraint的消融 以及paraphrase的消融 还有训练方式的消融（这里需要去讲RFT和RL）。

% (1) How do SFT and RL respectively affect model performance on 

The {\nickname} framework employs techniques to ensure the model can master adaptive reasoning from controllable and diverse synthetic data: (i) deploying the RLVR training strategy;
(ii) introducing the \textit{sanity check}, comprising Variable alignment (VA), Executable Code (EC), and Existence of Valid Query (EVQ);
(iii) incorporating the paraphrase to further enhance the diversity of query templates.

% 换个表述，在表格里的结果，显示了每个都有用
As shown in Table~\ref{tab:ablation_study},  each technique within the {\nickname} framework is essential to the final performance.
Notably, without EC, directly applying subsequent EVQ retains only $0.2\%$ data, which is insufficient to support training; consequently, no result is reported for this configuration.
Although this indicates that EVQ functionally contains EC, the rule-based EC is more computationally efficient than the model-based EVQ, by $218\times$ (as shown in Appendix~\ref{apx:cost}), and thus remains essential.
Besides, the result proves that paraphrase is complementary to perturbation of variable values, since it enhances the ability to understand different templates of specific problem-solving logic.

To further analysis the benefits of RLVR, we disentangle reasoning ability from computational ability.
Reasoning ability is evaluated by the accuracy of code generation. Computational ability is evaluated by providing the code and instructing the model to execute it (the full prompt is provided in Appendix~\ref{apx:prompt}).

As summarized in Table~\ref{tab:reason_compute}, training with standard RLVR (Standard-RLVR) represents a single-data setting where outcome rewards are inherently unreliable; spurious reasoning can easily stumble upon the correct answer, leading to a degradation in performance compared to ``Initial-SFT''. By contrast, the synthetic data generated by {\nickname} provides a rich, multi-data setting across varying numerical conditions. This diversity fundamentally prevents the model from overfitting to memorized numerical information. Consequently, even when simply applying Rejection Sampling Fine-Tuning on our perturbed data ({\nickname}-RFT), the model achieves overall performance gains over the baseline.

However, a closer look at the disentangled metrics reveals that ``{\nickname}-RFT'' improves computation while degrading reasoning relative to ``Initial-SFT''. This behavior is consistent with~\citet{chu2025sft}: RFT’s objective tends to memorize superficial features, thereby exacerbating overfitting to computation patterns rather than acquiring the underlying problem-solving logic.

\begin{table}[tbp]
    \centering
    \resizebox{\columnwidth}{!}{
    \begin{tabular}{c c c c}
    \toprule
    Method  & ORCA-{\nickname}-test & Reasoning & Computation \\
    \midrule
    Initial-SFT & 77.12 & 75.24 & 54.92 \\
    Standard-RLVR & 81.80 & 69.52 & 68.52 \\
    {\nickname}-RFT & 82.32 & 72.36 & 75.36 \\
    {\nickname} & \textbf{86.08} & \textbf{80.96} & \textbf{90.76} \\
    \bottomrule
    \end{tabular}}
    \caption{Evaluation of reasoning and computation ability on ORCA-{\nickname}.}
    \label{tab:reason_compute}
    \vspace{-15pt}
\end{table}

\begin{figure*}[t]
    \centering
    \includegraphics[width=\linewidth]{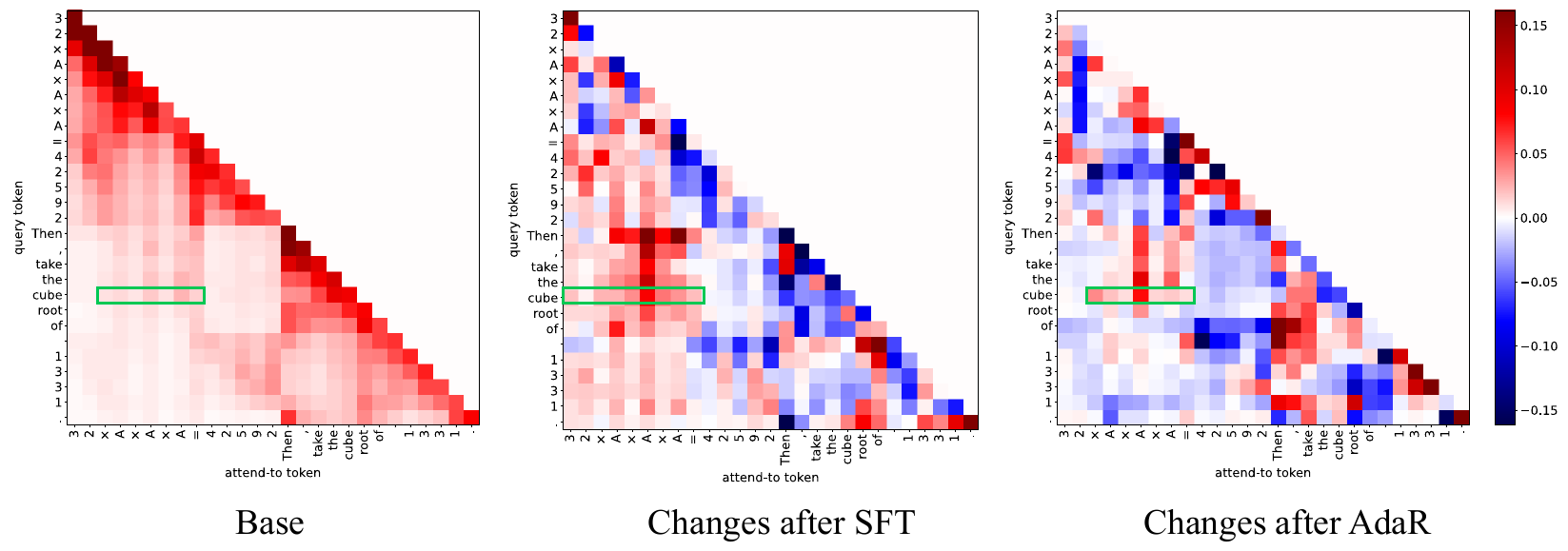}
    \caption{
    Attention heatmaps across different training stages. Green boxes highlight the attention weights associated with generating the token ``cube''.
    }
    \label{fig:attention}
\end{figure*}

\section{Analysis}

\vspace{-5pt}

In this subsection, we address the following Research Questions (RQs) regarding {\nickname}:
(1) Does training with {\nickname} enable the model to master adaptive reasoning?
% 不能说那个最关键，而是说影响程度分别有多大
(2) To what extent does the scaling along different dimensions (query template $T$, variable set $x$, and problem-solving logic $L$) influence the model’s performance respectively?
% 第五个按照讨论的逻辑说，在
(3) During the RLVR stage, when employing synthetic data whose templates are unseen by the target LLM, does {\nickname} continue to exhibit strong performance?
(4) To what extent does diversity in the computational difficulty levels of synthetic queries affect the model’s performance?
(5) Is {\nickname} applicable to Instruct models?
% 为了解答这些问题，我们进行了如下分析。就可以到正式分析了
To answer these RQs (excluding RQ5), we conduct the following analysis using Qwen2.5-MATH-7B as the base model. 
Because the AIME evaluation metric differs from those of the other out-of-domain benchmarks, we exclude AIME from the analysis for better reporting.

% To address \textbf{RQ1}, for a given query we evaluate the model's code generation capability as a proxy for reasoning. 
% We further probe the extent to which our method induces logic by analyzing the model's sensitivity to the order of correct CoT generated by itself.
% To address \textbf{RQ2}, we retain the above proxy for reasoning, and additionally use code execution capability as a proxy for computation, thereby enabling a decoupled analysis, thereby enabling a decoupled analysis.
% For \textbf{RQ3}, we vary the parameter $\alpha$ in controllable perturbation across magnitudes and examine the resulting performance changes. 
% For \textbf{RQ4}, we vary the scale for paraphrase the template , for numerical perturbation and for queries independently.
% To investigate \textbf{RQ5}, we evaluate the performance of {\nickname} when the shortcuts induced from SFT are removed.

\subsection{{\nickname} Enables Adaptive Reasoning}
\label{ana:master_logic}

% \paragraph{Enhancing the generalization and robustness.}
% In Section~\ref{sec:main_result_2}, we show that achieving generalization on out-of-domain test sets is indicative of the model's mastery of adaptive reasoning.
% Here we provide a more granular analysis on robustness of in-domain tasks.

% \paragraph{Enhancing the influence to logical order.}
\paragraph{Correcting attention towards templates.}
To concretely demonstrate how SFT induces spurious reasoning and how {\nickname} corrects it, we conduct an attention-based analysis (Figure~\ref{fig:attention}). Using the example from Figure~\ref{fig:example}, when generating the target reasoning token ``cube'', the model should ideally attend to the template ``$\times A \times A \times A=$'', which encodes the actual problem-solving logic $L$.
% It should not rely on the instantiated numerical values, such as $32$ and $42592$. 
Before SFT, the base model correctly focuses on the template. However, after SFT, the attention distribution becomes irregular, improperly elevating the importance of irrelevant numerical tokens like $32$. 
This confirms that SFT inadvertently encourages the model to associate reasoning steps with superficial numerical features. 
In contrast, after training with {\nickname}, the attention toward the template is substantially strengthened, while attention to specific variable-values diminishes. 
As the model learns to correct its attention and ground its predictions in templates rather than superficial numerical patterns, it also becomes strictly dependent on the correct sequence of logical steps---a phenomenon we quantitatively verify using our Influence to Logical Order (ILO) metric in Appendix~\ref{apx:ilo}.

\paragraph{Enhancing algebraic thinking.}
% 55%-90%怎么得到的说一下
Once the model stops memorizing specific numbers and aligns with the underlying problem-solving logic, it naturally develops stronger algebraic thinking.
We observe that, despite no code data being provided during training, the proportion of CoTs containing structural code snippets (e.g. ``... \texttt{B + H = 157 - (23 + 41). Substituting the value of H into the equation, we get} ...'') increases from $55\%$ to $90\%$ after training with {\nickname} (please refer to Appendix~\ref{apx:case} for more details), indicating the emergence of algebraic thinking, treating unknown and known variables on equal footing and solving queries via variable calculation.
We then likewise evaluate reasoning ability by measuring the accuracy of problem-solving code generation which engages algebraic thinking.
As shown in Table~\ref{tab:reason_compute}, although RLVR is employed in both, training on standard data (Standard-RLVR) decreases accuracy, whereas training on our synthetic data with queries which are semantically similar but instantiate different variables improves accuracy.
Furthermore, we provide a theoretical analysis in Appendix~\ref{apx:theory} to justify this phenomenon.

\begin{figure}[tbp]
  \centering
    \vspace{-6pt}
    \includegraphics[width=\linewidth]{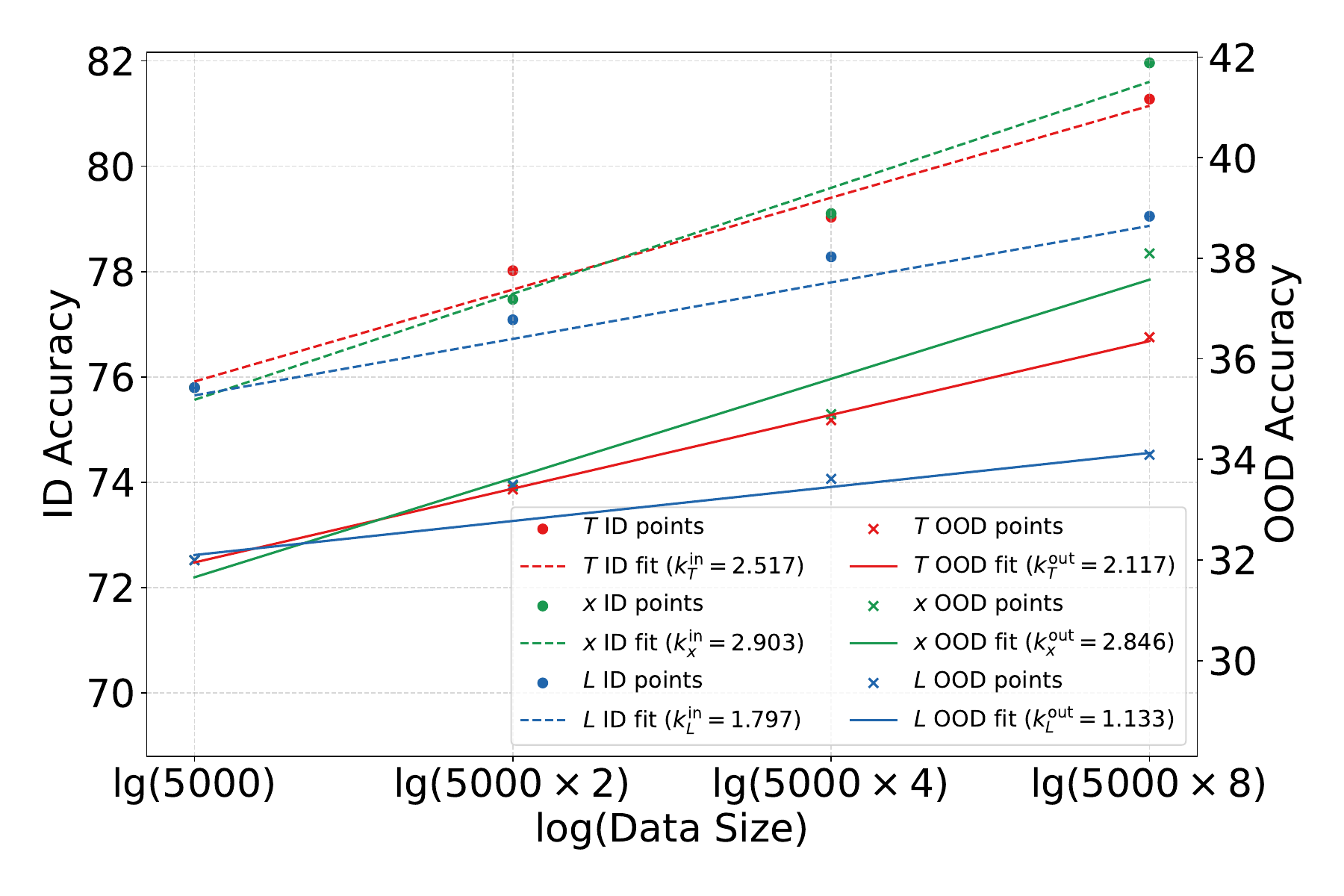}
    \vspace{-25pt}
    \caption{Performance of scaling.}
    \label{fig:scale_l}
    \vspace{-15pt}
\end{figure}

\subsection{Effect of Scaling Different Dimensions}
%  改写这个analysis，去讲多样性L x T多样性，去分别扰动，然后三个维度有主次之分
% gx：可能x和T的scaling是可比的，L和他们不可比，scaling不是一个维度
\label{sec:diversity} 

In {\nickname}, we modulate sampling frequency to scale data size along query template $T$ and values in variable set $x$.
In addition, we scale data size along problem-solving logic $L$ by subsampling, which regulates diversity in common scenarios
Accordingly, we evaluate performance trends when scaling $x$, $T$ or $L$ and show the result in Figure~\ref{fig:scale_l}. 
We observe steady improvements in model performance with increasing data size in both in-domain (ID) and out-of-domain (OOD) settings. 
To characterize marginal returns with scale, we fit the scaling curves with a log-linear approximation. 
The fitted coefficients satisfy $k_L < k_T < k_x$ throughout, indicating that scaling along the dimension $x$ yields the greatest marginal returns.
This is mainly because perturbing $T$ facilitates query comprehension, whereas perturbing $x$ is the essential driver of adaptive reasoning.
We attribute the limited effectiveness of scaling along $L$ to the simple subsampling, which causes data from same distribution constrains the control of diversity. We leave more comprehensive exploration of $L$ to future work.
 
% For a more comprehensive analysis, we also evaluate the effect of scaling along the problem-solving logic by increasing the data size of the seed dataset used for synthesis.
% The results are shown in Figure~\ref{fig:scale_r}.
% Overall, these three dimensions are complementary, but the cost of scaling $L$ is substantially higher; 
% Overall, these three dimensions are complementary, but the cost of scaling $L$ is substantially higher than $x$ and $T$.Therefore, in this paper we highlight the importance of scaling $L$ while leaving this point as future work.

\subsection{Effect of Seen Perturbed Queries}
\label{ana:learn_static_cot}

{\nickname} learns adaptive reasoning by comparing feedback on responses to perturbed queries that preserve the same problem-solving logic; hence, such queries should be included during RLVR. We identify two sources of comparison.

First, SFT may induce spurious reasoning by memorizing superficial features. When faced with perturbations of queries which are seen during SFT, the model produces unstable rollouts, yielding contrasting feedback. This is supported by Table~\ref{tab:remove_sft}, where training {\nickname} on perturbed data conditioned on SFT seed data outperforms ``Standard-RLVR'' trained on normal data without contrasts.

Second, contrast can also arise directly during RLVR: perturbed queries with shared logic but different values expose spurious reasoning through inconsistent feedback. To test this, we sample a non-overlapping 2.25K subset from ORCA-MATH and apply four rounds of \textit{controllable perturbation}, denoted ``{\nickname}-Lite$\times$4''. As shown in Table~\ref{tab:remove_sft}, this setting also outperforms ``Standard-RLVR'', further supporting our hypothesis.
% 不是解释这是一种限制，而是说这是符合理论的实验现象，侧面验证理论的合理性
% 理论是说需要有正负样本，那么如果在unseen数据上是否仍然能产生需要的正负样本，实验表明是可以的，因为这种fake reasoning会影响到unseen数据
% 在unseen的query使用我们的{\nickname}，仍然可以取得比较好的效果，这个由于SFT模型在unseen的query上一样会诱导出spurious reasoning

% Our preliminary experiments also show that such spurious reasoning is prevalent\footnote{In GSM-SYM main test set, $100$ symbolic templates are each instantiated $50$ times to form a group. 
% We compute the proportion of groups where Qwen2.5-Math-7B produces both correct answers (from spurious reasoning) and incorrect answers is $55\%$.}.

\subsection{Effect of Perturbation Magnitude}
\begin{figure}[hbp]
  \centering
        \includegraphics[width=0.8\linewidth]{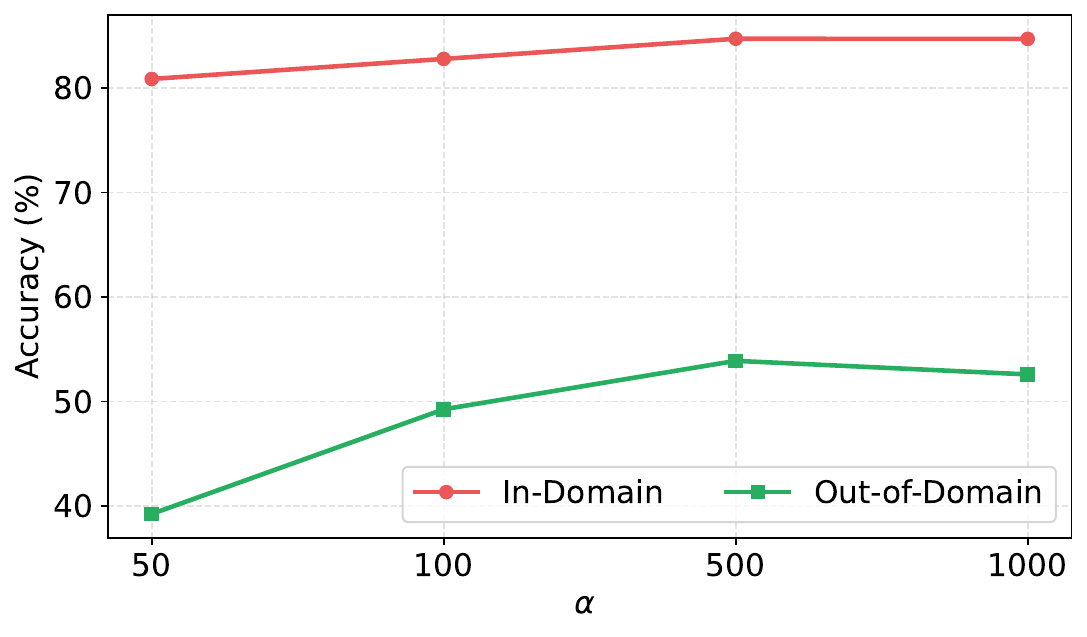}
  \vspace{-5pt}
    \caption{Influence of perturbation magnitude.}
    \label{fig:perturb_extent}
    \vspace{-15pt}
\end{figure}

We investigate the influence of perturbation magnitude and present the results in Figure~\ref{fig:perturb_extent}.
We observe that the performance increases with larger perturbation in variable values up to a point and then exhibits a slight decline.
Larger perturbation magnitude enables the model to learn previously unseen numerical calculations, and thereby improves performance.
However, an overly large $\alpha$ increases the likelihood that perturbed variable sets violate implicit inter-variable constraints, producing more invalid candidates before the \textit{sanity check}. If we suppose the success rate of sanity check is limited, these invalid candidates introduce additional noise, ultimately degrading model performance.
% Increasing the perturbation magnitude $\alpha$ in Equation~\ref{eq:perturb} increases the number of computational digits and, consequently, the computational difficulty.
% We control the diversity in computational difficulty by adjusting the perturbation magnitude $\alpha$ in the Equation~\ref{eq:perturb}.
% \begin{wrapfigure}{r}{0.5\textwidth}
%   \centering
%   \vspace{-\baselineskip} % 可选：把表格往上紧一点
%   \includegraphics[width=\linewidth]{fig/acc_vs_alpha.pdf}
%   \caption{Influence of perturbation magnitude.}
%   \label{fig:perturb_extent}
% \end{wrapfigure}

\begin{table}[t]
  \centering
  \resizebox{\linewidth}{!}{%
    \begin{tabular}{c c c c}
      \toprule
      Method & \# Samples & In-Domain & Out-of-Domain \\
      \midrule
      Standard-RLVR & 9K    & 77.82 & 37.20 \\
      {\nickname}-Lite$\times$4             & 2.25K $\times$ 4 & 83.42 & 49.90 \\
      {\nickname}             & 9K & 84.72 & 53.88 \\
      \bottomrule
    \end{tabular}%
  }
      \vspace{-5pt}
  \caption{Effect of seen perturbed queries.}
  \label{tab:remove_sft}
  \vspace{-10pt}
\end{table}

\subsection{{\nickname} is Applicable to Instruct Model}

\label{sec:instruct}

We further investigate using Qwen3-4B-Instruct-2507~\citep{qwen3technicalreport} as the base model.
{\nickname} still yields substantial performance gains (+5.86) and effectively mitigates robustness failures that persist under test-time scaling (+1.51). 
We additionally observe that the model's rollout time on seed dataset is shorter than that of \nickname{} ($\approx0.36\times$), suggesting that generating correct answers becomes more difficult when the model is exposed to mild numerical perturbations, revealing pronounced brittleness.
Taken together, these findings highlight a pronounced limitation in advanced LLMs: strong performance on conventional test sets often obscures fundamental deficiencies in generalization and robustness. 
In contrast, {\nickname} effectively exposes and mitigates such latent deficiencies, demonstrating its broad applicability.
Additional details are provided in Appendix~\ref{apx:instruct}.

% We further investigate whether {\nickname} can improve performance when ``Instruct'' is used as the base (noted as ``{\nickname}-Instruct''). 
% % 对比Instruct结果
% We find that when using Qwen2.5-MATH-7B as the base model, {\nickname} has already achieved performance comparable to its ``Instruct'' version (73.16 vs. 73.99), despite the latter being trained on 2.5M CoT data, approximately $3000\times$ more data than ours.
% Moreover, ``{\nickname}-Instruct'' achieves an additional improvement of 1.92 points, further demonstrating the broad applicability of {\nickname}. 
% Detailed results on each test set are provided in Appendix~\ref{apx:instruct}.

\section{Related Work}
% \cite{wangetal2023towards} 无效COT也可以取得和有效COT差不多的性能，但是基于prompt而不是训练
% \cite{prystawski2023why} 理论说明COT的意义
% \cite{feng2023towards} 理论说明COT的意义
% GSM SYM
%robustness of in-domain tasks~\citep{gsm_sym}
%generalization of out-of-domain tasks~\citep{jahin2025evaluating}.
\paragraph{Analysis of Robustness and Generalization.}
% CoT has shown success in enhancing reasoning performance, with theoretical explanations suggesting CoT enables the chaining of accurate local inferences to estimate relationships between unseen variables~\citep{prystawski2023why,feng2023towards}.
% between variables
% These training
% conditions enable the chaining of accurate local inferences to estimate relationships
% between variables that were not seen together in training. 
\citet{wangetal2023towards} discovered that even invalid CoT demonstrations achieve comparable performance to valid ones.
Similarly, we demonstrate that CoT arising from spurious reasoning is less sensitive to random variations in logical order.
LLMs exhibit brittleness when facing problem variations: the GSM-SYM benchmark \citep{gsm_sym} demonstrates poor robustness on in-domain tasks with altered numerical values, while \citet{jahin2025evaluating} shows limited generalization to out-of-domain problems, indicating models may memorize patterns rather than genuinely understand reasoning.
The generation of GSM-SYM relies on human effort. In contrast, {\nickname} is fully automatic.
% The generation of GSM-SYM relies on human effort, and therefore only a small amount of data can be generated for testing. In contrast, {\nickname} is fully automatic, produces high-quality data, and can be used for both training and testing.

\paragraph{Approaches to Improving Reasoning Ability.} 
Data synthesis has emerged as a promising direction \citep{wang2025comprehensivesurveydataaugmentation}. Prior work diversifies question templates via paraphrasing \citep{li2023mugglemath} or LLM-based generation and verification \citep{lu2024mathgenie}, but often preserves the underlying mathematical structures or relies on error-prone verification, limiting gains in adaptive reasoning. Our proposed notion of adaptive reasoning helps explain these limitations. Concurrently, \citet{liang2025beyond} synthesizes queries online from model responses and back-propagates training loss to guide data generation. Our method is complementary: incorporating the model under training into {\nickname} for template and code generation, combined with such online training, may further improve performance.

\section{Conclusion}
Robustness and generalization remain central challenges for LLMs when solving mathematical problems. 
We attribute these failures to spurious reasoning that relies on superficial features and encourage adaptive reasoning where models can establish correct correlation between templates and reasoning process. 
Therefore, we propose {\nickname}, a framework that enables adaptive reasoning and comprises a data synthesis component and a model training component. Experimental results demonstrate substantial performance improvements with a small amount of data on both in-domain and out-of-domain tasks. 
Further analyses indicate that {\nickname} indeed facilitates adaptive reasoning and is a scalable and broadly applicable framework.

\section*{Limitations}
Our framework is subject to several limitations. 
Firstly, although Section~\ref{sec:diversity} underscores the importance of scaling $L$, {\nickname} currently enhances only $T$ and $x$. 
We therefore identify scaling the diversity of $L$ as a promising direction for future work.
Secondly, {\nickname} requires the entire problem-solving process of each seed queries to be expressible as a fully executable program, and it assumes the presence of easily perturbed input variables. 
These requirements implies applying {\nickname} in more general settings (e.g., theorem proving) would require more sophisticated designs.
Thirdly, due to computational constraints, we do not investigate large reasoning models or models with larger parameter counts as base.

% \section*{Reproducibility Statement}
% We provide, in the supplementary materials, all settings necessary to reproduce our experimental results, including data synthesis scripts, training scripts and code, configuration files, and fixed random seeds. We will additionally release trained checkpoints to further facilitate reproduction.

\section*{Acknowledgments}
We would like to thank the anonymous reviewers for their insightful comments. Shujian Huang is the corresponding author. This work is supported by National Science Foundation of China (No. 62376116), the Fundamental Research Funds for the Central Universities (No. 2024300507), Fundamental and Interdisciplinary Disciplines Breakthrough Plan of the Ministry of Education of China (No. JYB2025XDXM118).

\bibliography{custom}

@article{cot,
  title={Chain-of-thought prompting elicits reasoning in large language models},
  author={Wei, Jason and Wang, Xuezhi and Schuurmans, Dale and Bosma, Maarten and Xia, Fei and Chi, Ed and Le, Quoc V and Zhou, Denny and others},
  journal={Advances in neural information processing systems},
  volume={35},
  pages={24824--24837},
  year={2022}
}

@inproceedings{wangetal2023towards,
    title = "Towards Understanding Chain-of-Thought Prompting: An Empirical Study of What Matters",
    author = "Wang, Boshi  and
      Min, Sewon  and
      Deng, Xiang  and
      Shen, Jiaming  and
      Wu, You  and
      Zettlemoyer, Luke  and
      Sun, Huan",
    editor = "Rogers, Anna  and
      Boyd-Graber, Jordan  and
      Okazaki, Naoaki",
    booktitle = "Proceedings of the 61st Annual Meeting of the Association for Computational Linguistics (Volume 1: Long Papers)",
    month = jul,
    year = "2023",
    address = "Toronto, Canada",
    publisher = "Association for Computational Linguistics",
    url = "https://aclanthology.org/2023.acl-long.153/",
    doi = "10.18653/v1/2023.acl-long.153",
    pages = "2717--2739"
}

@inproceedings{math_survey,
  title={Towards Reasoning in Large Language Models: A Survey},
  author={Huang, Jie and Chang, Kevin Chen-Chuan},
  booktitle={61st Annual Meeting of the Association for Computational Linguistics, ACL 2023},
  pages={1049--1065},
  year={2023},
  organization={Association for Computational Linguistics (ACL)}
}

@article{jahin2025evaluating,
  title={Evaluating Mathematical Reasoning Across Large Language Models: A Fine-Grained Approach},
  author={Jahin, Afrar and Zidan, Arif Hassan and Zhang, Wei and Bao, Yu and Liu, Tianming},
  journal={arXiv preprint arXiv:2503.10573},
  year={2025}
}

@article{gsm_sym,
  title={Gsm-symbolic: Understanding the limitations of mathematical reasoning in large language models},
  author={Mirzadeh, Iman and Alizadeh, Keivan and Shahrokhi, Hooman and Tuzel, Oncel and Bengio, Samy and Farajtabar, Mehrdad},
  journal={arXiv preprint arXiv:2410.05229},
  year={2024}
}

@article{li2023mugglemath,
  title={Mugglemath: Assessing the impact of query and response augmentation on math reasoning},
  author={Li, Chengpeng and Yuan, Zheng and Yuan, Hongyi and Dong, Guanting and Lu, Keming and Wu, Jiancan and Tan, Chuanqi and Wang, Xiang and Zhou, Chang},
  journal={arXiv preprint arXiv:2310.05506},
  year={2023}
}

@inproceedings{lu2024mathgenie,
  title={Mathgenie: Generating synthetic data with question back-translation for enhancing mathematical reasoning of llms},
  author={Lu, Zimu and Zhou, Aojun and Ren, Houxing and Wang, Ke and Shi, Weikang and Pan, Junting and Zhan, Mingjie and Li, Hongsheng},
  booktitle={Proceedings of the 62nd Annual Meeting of the Association for Computational Linguistics (Volume 1: Long Papers)},
  pages={2732--2747},
  year={2024}
}

@article{gsm8k,
  title={Training verifiers to solve math word problems},
  author={Cobbe, Karl and Kosaraju, Vineet and Bavarian, Mohammad and Chen, Mark and Jun, Heewoo and Kaiser, Lukasz and Plappert, Matthias and Tworek, Jerry and Hilton, Jacob and Nakano, Reiichiro and others},
  journal={arXiv preprint arXiv:2110.14168},
  year={2021}
}

@article{orca-math,
  title={Orca-math: Unlocking the potential of slms in grade school math},
  author={Mitra, Arindam and Khanpour, Hamed and Rosset, Corby and Awadallah, Ahmed},
  journal={arXiv preprint arXiv:2402.14830},
  year={2024}
}

@article{MATH,
  title={Measuring mathematical problem solving with the math dataset},
  author={Hendrycks, Dan and Burns, Collin and Kadavath, Saurav and Arora, Akul and Basart, Steven and Tang, Eric and Song, Dawn and Steinhardt, Jacob},
  journal={arXiv preprint arXiv:2103.03874},
  year={2021}
}

@article{colleage_math,
  title={Mathscale: Scaling instruction tuning for mathematical reasoning},
  author={Tang, Zhengyang and Zhang, Xingxing and Wang, Benyou and Wei, Furu},
  journal={arXiv preprint arXiv:2403.02884},
  year={2024}
}

@article{theoremqa,
  title={Theoremqa: A theorem-driven question answering dataset},
  author={Chen, Wenhu and Yin, Ming and Ku, Max and Lu, Pan and Wan, Yixin and Ma, Xueguang and Xu, Jianyu and Wang, Xinyi and Xia, Tony},
  journal={arXiv preprint arXiv:2305.12524},
  year={2023}
}

@article{yu2025dapo,
  title={Dapo: An open-source llm reinforcement learning system at scale},
  author={Yu, Qiying and Zhang, Zheng and Zhu, Ruofei and Yuan, Yufeng and Zuo, Xiaochen and Yue, Yu and Dai, Weinan and Fan, Tiantian and Liu, Gaohong and Liu, Lingjun and others},
  journal={arXiv preprint arXiv:2503.14476},
  year={2025}
}

@inproceedings{llamafactory,
  title={LlamaFactory: Unified Efficient Fine-Tuning of 100+ Language Models},
  author={Yaowei Zheng and Richong Zhang and Junhao Zhang and Yanhan Ye and Zheyan Luo and Zhangchi Feng and Yongqiang Ma},
  booktitle={Proceedings of the 62nd Annual Meeting of the Association for Computational Linguistics (Volume 3: System Demonstrations)},
  address={Bangkok, Thailand},
  publisher={Association for Computational Linguistics},
  year={2024},
  url={http://arxiv.org/abs/2403.13372}
}

@article{yu2023metamath,
  title={Metamath: Bootstrap your own mathematical questions for large language models},
  author={Yu, Longhui and Jiang, Weisen and Shi, Han and Yu, Jincheng and Liu, Zhengying and Zhang, Yu and Kwok, James T and Li, Zhenguo and Weller, Adrian and Liu, Weiyang},
  journal={arXiv preprint arXiv:2309.12284},
  year={2023}
}

@article{yang2024qwen2,
  title={Qwen2. 5-math technical report: Toward mathematical expert model via self-improvement},
  author={Yang, An and Zhang, Beichen and Hui, Binyuan and Gao, Bofei and Yu, Bowen and Li, Chengpeng and Liu, Dayiheng and Tu, Jianhong and Zhou, Jingren and Lin, Junyang and others},
  journal={arXiv preprint arXiv:2409.12122},
  year={2024}
}

@article{grattafiori2024llama,
  title={The llama 3 herd of models},
  author={Grattafiori, Aaron and Dubey, Abhimanyu and Jauhri, Abhinav and Pandey, Abhinav and Kadian, Abhishek and Al-Dahle, Ahmad and Letman, Aiesha and Mathur, Akhil and Schelten, Alan and Vaughan, Alex and others},
  journal={arXiv preprint arXiv:2407.21783},
  year={2024}
}

@misc{wang2025comprehensivesurveydataaugmentation,
      title={A Comprehensive Survey on Data Augmentation}, 
      author={Zaitian Wang and Pengfei Wang and Kunpeng Liu and Pengyang Wang and Yanjie Fu and Chang-Tien Lu and Charu C. Aggarwal and Jian Pei and Yuanchun Zhou},
      year={2025},
      eprint={2405.09591},
      archivePrefix={arXiv},
      primaryClass={cs.LG},
      url={https://arxiv.org/abs/2405.09591}, 
}

@article{wei2022chain,
  title={Chain-of-thought prompting elicits reasoning in large language models},
  author={Wei, Jason and Wang, Xuezhi and Schuurmans, Dale and Bosma, Maarten and Xia, Fei and Chi, Ed and Le, Quoc V and Zhou, Denny and others},
  journal={Advances in neural information processing systems},
  volume={35},
  pages={24824--24837},
  year={2022}
}

@article{shao2024deepseekmath,
  title={Deepseekmath: Pushing the limits of mathematical reasoning in open language models},
  author={Shao, Zhihong and Wang, Peiyi and Zhu, Qihao and Xu, Runxin and Song, Junxiao and Bi, Xiao and Zhang, Haowei and Zhang, Mingchuan and Li, YK and Wu, Yang and others},
  journal={arXiv preprint arXiv:2402.03300},
  year={2024}
}

@article{liu2021pretrainpromptpredictsystematic,
  title={Pre-train, prompt, and predict: A systematic survey of prompting methods in natural language processing},
  author={Liu, Pengfei and Yuan, Weizhe and Fu, Jinlan and Jiang, Zhengbao and Hayashi, Hiroaki and Neubig, Graham},
  journal={ACM computing surveys},
  volume={55},
  number={9},
  pages={1--35},
  year={2023},
  publisher={ACM New York, NY}
}

@article{brown2020language,
  title={Language models are few-shot learners},
  author={Brown, Tom and Mann, Benjamin and Ryder, Nick and Subbiah, Melanie and Kaplan, Jared D and Dhariwal, Prafulla and Neelakantan, Arvind and Shyam, Pranav and Sastry, Girish and Askell, Amanda and others},
  journal={Advances in neural information processing systems},
  volume={33},
  pages={1877--1901},
  year={2020}
}

@article{guo2025deepseek,
  title={Deepseek-r1: Incentivizing reasoning capability in llms via reinforcement learning},
  author={Guo, Daya and Yang, Dejian and Zhang, Haowei and Song, Junxiao and Zhang, Ruoyu and Xu, Runxin and Zhu, Qihao and Ma, Shirong and Wang, Peiyi and Bi, Xiao and others},
  journal={arXiv preprint arXiv:2501.12948},
  year={2025}
}

@inproceedings{chu2023navigate,
  title={Navigate through enigmatic labyrinth a survey of chain of thought reasoning: Advances, frontiers and future},
  author={Chu, Zheng and Chen, Jingchang and Chen, Qianglong and Yu, Weijiang and He, Tao and Wang, Haotian and Peng, Weihua and Liu, Ming and Qin, Bing and Liu, Ting},
  booktitle={Proceedings of the 62nd Annual Meeting of the Association for Computational Linguistics (Volume 1: Long Papers)},
  pages={1173--1203},
  year={2024}
}

@article{qwen3technicalreport,
  title={Qwen3 technical report},
  author={Yang, An and Li, Anfeng and Yang, Baosong and Zhang, Beichen and Hui, Binyuan and Zheng, Bo and Yu, Bowen and Gao, Chang and Huang, Chengen and Lv, Chenxu and others},
  journal={arXiv preprint arXiv:2505.09388},
  year={2025}
}

@article{chu2025sft,
  title={Sft memorizes, rl generalizes: A comparative study of foundation model post-training},
  author={Chu, Tianzhe and Zhai, Yuexiang and Yang, Jihan and Tong, Shengbang and Xie, Saining and Schuurmans, Dale and Le, Quoc V and Levine, Sergey and Ma, Yi},
  journal={arXiv preprint arXiv:2501.17161},
  year={2025}
}

@misc{yuan2023scaling,
      title={Scaling Relationship on Learning Mathematical Reasoning with Large Language Models}, 
      author={Zheng Yuan and Hongyi Yuan and Chengpeng Li and Guanting Dong and Keming Lu and Chuanqi Tan and Chang Zhou and Jingren Zhou},
      year={2023},
      eprint={2308.01825},
      archivePrefix={arXiv},
      primaryClass={cs.CL}
}

@article{sheng2024hybridflow,
  title   = {HybridFlow: A Flexible and Efficient RLHF Framework},
  author  = {Guangming Sheng and Chi Zhang and Zilingfeng Ye and Xibin Wu and Wang Zhang and Ru Zhang and Yanghua Peng and Haibin Lin and Chuan Wu},
  year    = {2024},
  journal = {arXiv preprint arXiv: 2409.19256}
}

@article{loshchilov2017decoupled,
  title={Decoupled weight decay regularization},
  author={Loshchilov, Ilya and Hutter, Frank},
  journal={arXiv preprint arXiv:1711.05101},
  year={2017}
}

@misc{tong2024dartmath,
  title={DART-Math: Difficulty-Aware Rejection Tuning for Mathematical Problem-Solving},
  author={Yuxuan Tong and Xiwen Zhang and Rui Wang and Ruidong Wu and Junxian He},
  year={2024},
  eprint={2407.13690},
  archivePrefix={arXiv},
  primaryClass={cs.CL},
  url={https://arxiv.org/abs/2407.13690},
}

@article{kang2025quagmires,
  title={Quagmires in SFT-RL Post-Training: When High SFT Scores Mislead and What to Use Instead},
  author={Kang, Feiyang and Kuchnik, Michael and Padthe, Karthik and Vlastelica, Marin and Jia, Ruoxi and Wu, Carole-Jean and Ardalani, Newsha},
  journal={arXiv preprint arXiv:2510.01624},
  year={2025}
}

@inproceedings{wen2025light,
  title={Light-r1: Curriculum sft, dpo and rl for long cot from scratch and beyond},
  author={Wen, Liang and Cai, Yunke and Xiao, Fenrui and He, Xin and An, Qi and Duan, Zhenyu and Du, Yimin and Liu, Junchen and Tanglifu, Tanglifu and Lv, Xiaowei and others},
  booktitle={Proceedings of the 63rd Annual Meeting of the Association for Computational Linguistics (Volume 6: Industry Track)},
  pages={318--327},
  year={2025}
}

@article{teng2025atom,
  title={Atom of thoughts for markov llm test-time scaling},
  author={Teng, Fengwei and Shi, Quan and Yu, Zhaoyang and Zhang, Jiayi and Luo, Yuyu and Wu, Chenglin and Guo, Zhijiang},
  journal={arXiv preprint arXiv:2502.12018},
  year={2025}
}

@article{liang2025beyond,
  title={Beyond pass@ 1: Self-play with variational problem synthesis sustains rlvr},
  author={Liang, Xiao and Li, Zhongzhi and Gong, Yeyun and Shen, Yelong and Wu, Ying Nian and Guo, Zhijiang and Chen, Weizhu},
  journal={arXiv preprint arXiv:2508.14029},
  year={2025}
}

@article{huang2025survey,
  title={A survey on hallucination in large language models: Principles, taxonomy, challenges, and open questions},
  author={Huang, Lei and Yu, Weijiang and Ma, Weitao and Zhong, Weihong and Feng, Zhangyin and Wang, Haotian and Chen, Qianglong and Peng, Weihua and Feng, Xiaocheng and Qin, Bing and others},
  journal={ACM Transactions on Information Systems},
  volume={43},
  number={2},
  pages={1--55},
  year={2025},
  publisher={ACM New York, NY}
}

@article{Arjovsky2019InvariantRM,
  title={Invariant Risk Minimization},
  author={Mart{\'i}n Arjovsky and L{\'e}on Bottou and Ishaan Gulrajani and David Lopez-Paz},
  journal={ArXiv},
  year={2019},
  volume={abs/1907.02893},
  url={https://api.semanticscholar.org/CorpusID:195820364}
}

\appendix
\section{Appendix}

\subsection{The use of LLMs}
We use LLMs (e.g., GPT-5) only to polish writing. More specifically, their application focuses on two key areas: correcting grammatical errors and suggesting more appropriate word choices to enhance expression. Additionally, we conduct a thorough double check of all content refined by LLMs. This verification process is critical to preventing the inclusion of harmful information and ensuring the overall accuracy, reliability, and appropriateness of the final content.

\subsection{Training Objective}
\label{apx:obj}
\subsubsection{SFT \& RFT}
% SFT讲pi\theta的定义
Supervised fine-tuning (SFT) is a mainstream post-training strategy for eliciting step-by-step reasoning. 
Let $\pi_\theta$ denote a policy over responses given queries, parameterized by $\theta$. 
Given a dataset $\mathcal{D}$ of query–response pairs $(q, r)$, SFT minimizes the negative log-likelihood objective
\begin{equation}
    \mathcal{L}_{\text{SFT}}(\theta) = -\mathbb{E}_{(q,r)\sim\mathcal{D}}\big[\log \pi_\theta(r\mid q)\big],
\end{equation}
Reject sampling Fine-Tuning (RFT) constructs an SFT dataset by sampling from the model to be trained and retaining high-scoring outputs.

\subsubsection{RLVR}
Reinforcement Learning with Verifiable Reward (RLVR) is a reinforcement learning method that saves video memory usage.
Given a query \(q\), the model samples \(r\sim\pi_\theta(\cdot\mid q)\) and a verifier \(v(q,r)\in[0,1]\) evaluates it. RLVR maximizes
\begin{equation} J(\theta)=\mathbb{E}_{q\sim\mathcal{D}, r\sim \pi(\cdot\mid q)}[v(q, r)]
\end{equation}
In mathematical reasoning, \(v\) typically checks whether the predicted answer exactly matches the gold answer. 

\subsection{Prompts}
\label{apx:prompt}
We show the prompts used for generating the query template and problem-solving code in Prompt \ref{box:prompt1}, Solution Existence in Prompt \ref{box:prompt2}, code generation which evaluates reasoning ability in Prompt \ref{box:prompt3}, code execution which evaluates computational ability in Prompt \ref{box:prompt4}.

\subsection{General Settings}
\subsubsection{Data Synthesis}
\label{apx:data_syn}

Given that most publicly available math word problem datasets are constructed based on GSM8K~\citep{gsm8k}, and considering that existing LLMs may have been trained on substantial variations of this dataset, which alleviates to some extent the model's shortcut learning issues, we have opted for a newer and larger dataset ORCA-MATH~\citep{orca-math}.
We sample the data for synthesis by uniformly selecting using 42 as the random seed.

We use Qwen2.5-72B for data synthesis, including problem-solving codes, query templates, the existence of valid queries, and implementations of other baselines.
The temperature is set to 0.7, the top p is set to 0.95, and the maximum generation length is 4096 tokens.

In addition, we experiment with using Qwen2.5-7B for data synthesis under the same parameter settings, with the results reported in the Table~\ref{tab:data_syn}. Although employing a weaker model for data synthesis leads to a reduction in overall model performance, we observe that by scaling data size along the variable value $x$, the performance gap can be reduced from 8.14 to 3.09.

\subsubsection{Training}
\label{apx:training}
All baselines are first trained with the same SFT procedure, after which RLVR training is performed using their respective data.
For SFT, we use LLaMA-Factory~\citep{llamafactory}. 
All models are fine-tuned for 3 epochs with a batch size of 128 on 4 NVIDIA A100 GPUs. 
The peak learning rate is 1e-5 with a linear warm-up over the first 3\% of training steps, followed by cosine decay to a minimum of 1e-7.
The maximum generation length is set to 4096 tokens.
For DAPO, we use veRL~\citep{sheng2024hybridflow}, adopting the training setup of \citet{yu2025dapo}.
We utilize the AdamW optimizer~\citep{loshchilov2017decoupled} with a constant learning rate of 1e-6 and a linear warm-up over 20 rollout steps. For rollout, the prompt batch size is 256 and we sample 16 responses per prompt. For training, the mini-batch size is 32, corresponding to 128 gradient updates per rollout step.
Overall, SFT training requires approximately 12 hours and RL training requires approximately 2 days.

\subsubsection{Evaluation}
\label{apx:eva}

We compare {\nickname} with baselines on the following 7 benchmarks:
\begin{itemize}
    \item \textbf{GSM8K} [MIT]~\citep{gsm8k}: The test set comprises 1,319 high-quality grade-school mathematics word problems, each requiring between 2 and 8 reasoning steps. 
    \item \textbf{ORCA-{\nickname}-test}: The test set consists of 2,500 high-quality grade-school mathematics word problems synthesized by {\nickname}. Seed problems are selected from ORCA [MIT License]~\citep{orca-math} and subjected to 1-4 numerical perturbations to assess models’ robustness in mathematical reasoning.
    \item \textbf{GSM-SYM} [CC-BY-4.0]~\citep{gsm_sym}: The test set includes three subsets—\emph{main}, \emph{p1}, and \emph{p2}—with 5{,}000, 5{,}000, and 2{,}500 instances, respectively. Starting from the GSM8K test problems, a symbolic template is constructed for each problem; each template yields 50 instances. The subsets augment the original logical structure by adding 0 (\emph{main}), 1 (\emph{p1}), or 2 (\emph{p2}) additional reasoning clauses, thereby enabling a more rigorous evaluation of mathematical reasoning robustness.
    \item \textbf{MATH} [MIT]~\citep{MATH}: The test set comprises 5,000 problems drawn from high-school mathematics competitions. Problems are categorized into seven types (Prealgebra, Intermediate Algebra, Algebra, Precalculus, Geometry, Counting \& Probability, and Number Theory) and five difficulty levels.
    \item \textbf{CollegeMath}~\citep{colleage_math}: The test set contains 2,818 college-level problems curated from nine college-level mathematics textbooks, covering seven key disciplines: Algebra, Precalculus, Calculus, VectorCalculus, Probability, LinearAlgebra, and Differential Equations.
    \item \textbf{TheoremQA}~\citep{theoremqa}: A theorem-driven question-answering benchmark containing 800 problems grounded in 350 theorems, designed to evaluate LLMs’ ability to apply domain-specific theorems across Mathematics, Physics, Electrical Engineering, Computer Science, and Finance.
    \item \textbf{AIME 2025}: A test set containing 30 problems from the 2025 American Invitational Mathematics Examination (AIME), curated to evaluate LLMs on challenging, Olympiad-level high-school mathematics across Algebra, Geometry, Number Theory, and Combinatorics.
\end{itemize}

We adopt the evaluation pipeline of~\citet{tong2024dartmath} with some modifications. 
Unless otherwise noted, we set the sampling temperature to 0.7 and the nucleus parameter top\_p to 0.9.

Regarding evaluation metrics, pass@1 is defined as the accuracy of the first sampled output, whereas avg@32 is the mean accuracy computed over $32$ sampled outputs.
As shown in Table~\ref{tab:pass_avg}, when the test set is sufficiently large (approximately 1,000 samples), the performance differences between pass@1 and avg@32 remain within 1 percentage point and the avg@32 results fully consistent with the conclusions drawn from the avg@1 results.

\begin{table}[h]
\centering
\resizebox{\columnwidth}{!}{
\begin{tabular}{lccc}
\hline
\textbf{Method} & \textbf{GSM8K} & \textbf{GSM-SYM-main} & \textbf{MATH} \\
\hline
Initial SFT     & 81.43 / 80.42 & 74.26 / 73.89 & 42.84 / 42.57 \\
Standard-RLVR   & 86.96 / 86.88 & 81.34 / 81.51 & 61.71 / 61.29 \\
MetaMATH        & 84.61 / 83.80 & 79.78 / 78.88 & 66.92 / 66.29 \\
\nickname{}            & 91.81 / 92.32 & 89.72 / 89.39 & 75.90 / 76.00 \\
\hline
\end{tabular}}
\caption{pass@1 / avg@32 results across different benchmarks.}
\label{tab:pass_avg}
\end{table}

\subsection{Analysis on Influence to Logical Order}
\label{apx:ilo}

As discussed in Section~\ref{ana:master_logic}, spurious reasoning relies on superficial features and can often derive answers without adhering to a strict logical sequence. By contrast, genuine adaptive reasoning requires strict adherence to a correct logical order to derive answers step by step.

To quantify the model's reliance on structural logic versus superficial patterns, we introduce a metric termed Influence to Logical Order (ILO). ILO measures the relative rate of change in the perplexity (PPL) of the correct answer when the sentence order of the corresponding Chain-of-Thought (CoT) is shuffled:
\begin{equation}
\label{eq:ilo}
\text{ILO}=\frac{1}{n}\sum_{i=1}^n\frac{\big|\text{PPL}(y \mid q, z) - \text{PPL}(y \mid q, R^i(z))\big|}{\text{PPL}(y \mid q, z)},
\end{equation}
where $q$ denotes the query, $z$ denotes the CoT, $y$ denotes the answer, $R$ denotes a random permutation function operating at the sentence level, and $n$ denotes the number of random permutations (set to $5$ to minimize variance).

We observe that the ILO of CoTs that fail to generate correct answers across all variable perturbations within a specific template is significantly lower than that of CoTs capable of producing universally correct answers across all such perturbations ($114.24\%$ vs. $221.87\%$). This substantiates that spurious reasoning (characterized by a lower ILO metric) is less sensitive to logical sequence and is more likely to induce generalization deficiencies, whereas adaptive reasoning strictly depends on logical order.

Consequently, we utilize ILO to compare the reasoning paradigms before and after applying our method. We show that training with {\nickname} significantly improves the adaptive reasoning capability over the baseline ``Initial SFT'' ($150.49\%$ vs. $119.22\%$), demonstrating that our framework successfully compels the model to rely on rigorous logical sequencing rather than superficial feature matching.

\subsection{Theoretical Analysis}
\label{apx:theory}

We formalize a mathematical reasoning problem as a triplet $(T, x, L)$, where $T$ represents the underlying query template (i.e., the logical structure), $x$ denotes the instance-specific numerical variables, and $L$ is the invariant problem-solving logic. A specific question is generated as:
\[
q = T(x)
\]
and the corresponding correct reasoning chain is denoted as $L(x)$. The model $\pi$ is trained to predict $L(x)$ given the query $q$.

\paragraph{Spurious Correlations \& The Limit of Varying Templates.}
In standard training distributions $\mathcal{D}$ (e.g., standard RLVR or SFT on static datasets), models often overfit to specific numbers. Even if $T$ are extensively varied, if the entropy of the numerical variables $H(x)$ remains artificially low (e.g., repeating the same integers across thousands of paraphrased texts), superficial features $s(x)$ (e.g., specific variable values) remain strongly coupled with the logic. Formally, the conditional mutual information is strictly positive:
\[
I_{\mathcal{D}}(s(x); L(x) \mid T) > 0
\]
Consequently, a risk-minimizing model can artificially inflate the conditional probability $P(L \mid q)$ by exploiting $s(x)$ as a shortcut. This leads to severe overfitting, where the model memorizes the mapping between specific numbers and the reasoning paths.

\paragraph{De-correlation via Variable Perturbation.}
To mitigate this, we construct a variable-perturbed distribution $\mathcal{D}_T$ by sampling diverse and randomized variables $x$ for each template $T$. Logically, the specific numerical values should be independent of the overarching problem-solving logic. By extensively perturbing $x$, we strictly decorrelate the superficial features from the logic. Thus, under the perturbed distribution, the conditional mutual information drops to zero:
\[
I_{\mathcal{D}_T}(s(x); L(x) \mid T) = 0
\]
In this setting, superficial numerical features are no longer predictive of the correct reasoning path within any given template.

\paragraph{Risk Minimization and Invariant Learning.}
To mitigate this, our variable-perturbed distribution $\mathcal{D}_T$ strictly decorrelates superficial features from the logic by extensively randomizing $x$, enforcing $I_{\mathcal{D}_T}(s(x); L(x) \mid T) = 0$. From the perspective of Invariant Risk Minimization (IRM)~\citep{Arjovsky2019InvariantRM}, a suboptimal shortcut predictor $\pi_{\text{shortcut}}(q) = f(s(x))$ evaluated on $\mathcal{D}_T$ reduces to random guessing, bounded by a positive Bayes error:
$$
\inf_{\pi_{\text{shortcut}}} R_{\mathcal{D}_T}(\pi_{\text{shortcut}}(q)) \ge \epsilon_{\text{random}} > 0
$$
Conversely, since $q = T(x)$ inherently contains sufficient information to recover $x$, an invariant logic predictor that extracts $x$ and applies the generalized mapping $L(x)$ achieves zero theoretical risk:
$$
\inf_{\pi_{\text{logic}}} R_{\mathcal{D}_T}(\pi_{\text{logic}}(q)) = 0
$$
Thus, empirical risk minimization on $\mathcal{D}_T$ forces the model to discard heuristic shortcuts and approximate the true invariant mapping $\pi^*(q) = \mathbb{E}[L(x) \mid q]$, compelling it to learn algebraic reasoning rather than memorization.

\subsection{Structural Code Snippet}
\label{apx:case}

\begin{figure*}
    \centering
     \includegraphics[width=\linewidth]{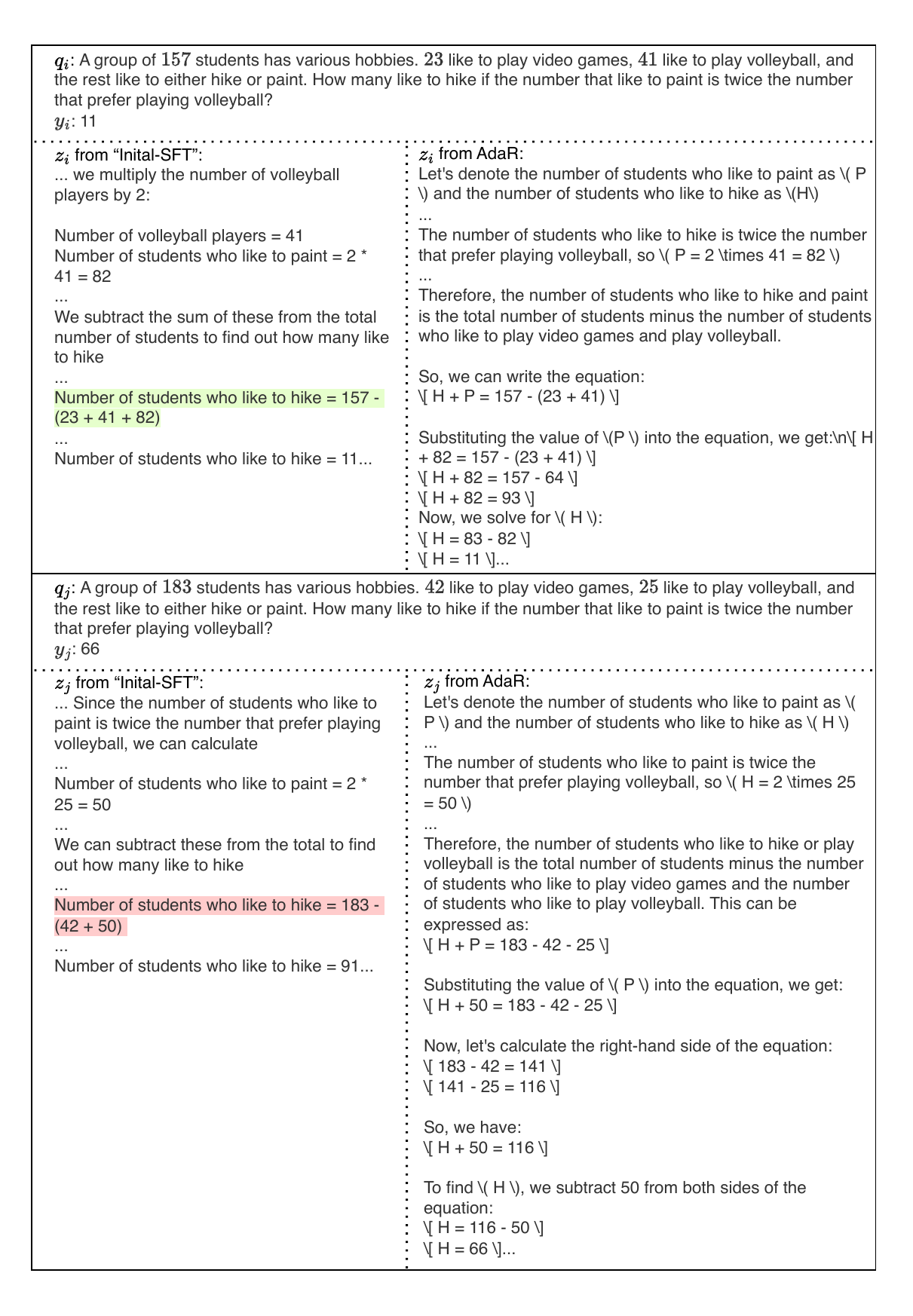}
    \caption{The case study of structural text in outputs. The green background indicates the correct reasoning step. The red background indicates the wrong reasoning step.}
    \label{fig:case_study}
\end{figure*}

To calculate the frequency of structural code snippet in the responses, we inspect 20 responses to queries from GSM-SYM. We present cases of structural code snippets from ``Initial-SFT'' and {\nickname} in Figure~\ref{fig:case_study}.

\subsection{Generation Cost}
\label{apx:cost}

The generation costs for all aspects included in the \textit{sanity check} are reported in Table \ref{tab:sanity_check}. 
This experiment is conducted on a single compute node.

\subsection{Performance on Instruct Models}
\label{apx:instruct}
\paragraph{Dataset.} To evaluate whether \nickname{} remains effective on more challenging datasets, beyond the grade-school mathematics problems used in earlier experiments (i.e., ORCA dataset), we adopt the Light-R1 dataset~\citep{wen2025light}. 
From this dataset, we sample 7.5K seed data and apply numerical perturbations 1–4 times, resulting in a 17K-instance training set, denoted as ``R1-\nickname{}-train''.
To perform a fine-grained analysis of whether \nickname{} improves model robustness and to avoid potential of data leakage, we construct two test sets, each containing 1K instances: ``seen'', whose query templates appear in ``R1-\nickname{}-train'', and ``unseen'', whose templates do not appear in the ``R1-\nickname{}-train''.

\paragraph{Baselines.} In this experiment, we introduce two strong baselines: ``Qwen3-4B-Instruct-2507'' and the test-time-scaling-enabled ``Qwen3-4B-Thinking-2507''. Because the Instruct model already possesses strong mathematical reasoning capabilities, we skip the SFT stage and directly perform RLVR training.
The model trained using RLVR on the 7.5K seed data is referred to as ``Standard-RLVR-Instruct'', while the model trained with \nickname{} synthetic data is denoted ``\nickname{}-Instruct''.

Table~\ref{tab:instruct_result} shows the full results.

% Table~\ref{tab:instruct_result} presents the performance of the Instruct model alongside the model further trained with {\nickname}. 

\subsection{Training Dynamic Analysis}

\begin{figure}[b]
    \centering
    \includegraphics[width=\linewidth]{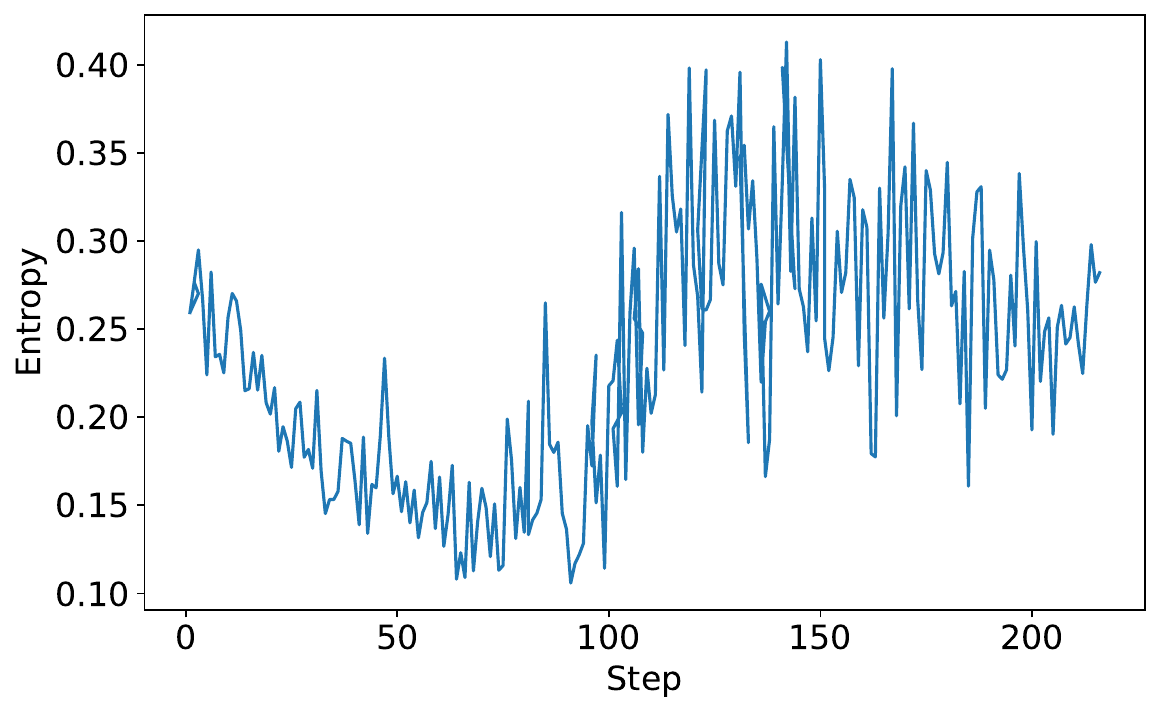}
    \caption{Entropy of actor model.}
    \label{fig:entropy}
\end{figure}

\begin{figure}[b]
    \centering
    \includegraphics[width=\linewidth]{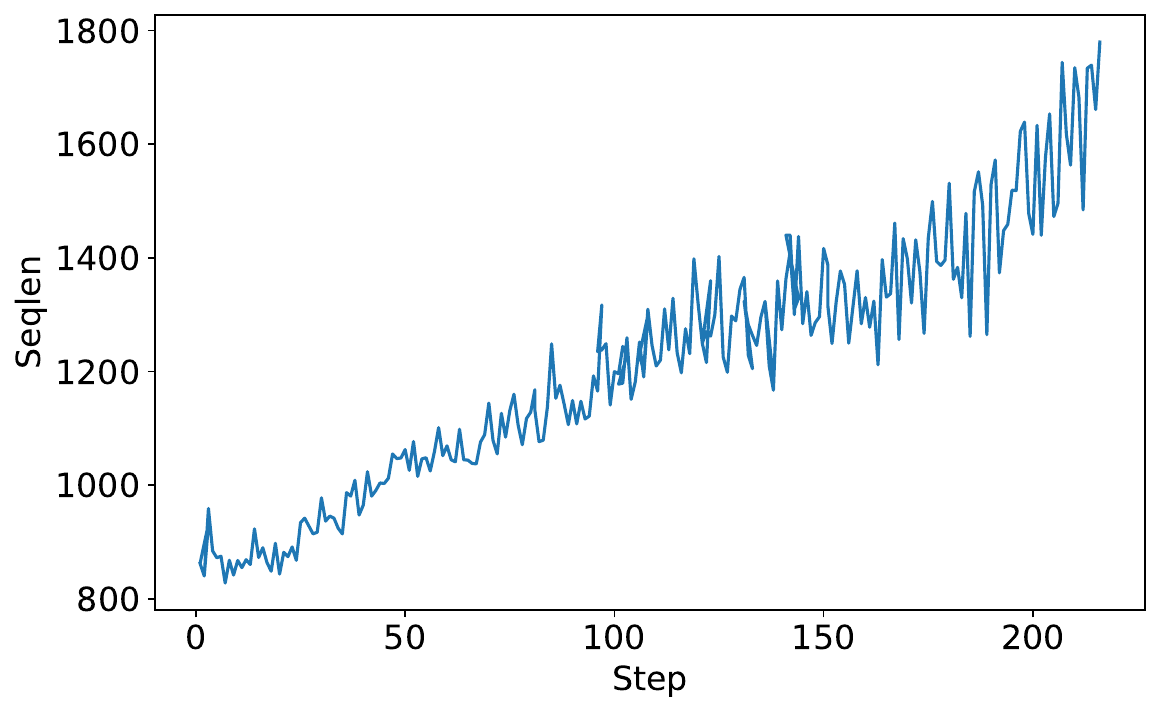}
    \caption{Generated sequence length of actor model.}
    \label{fig:seqlen}
\end{figure}

As shown in Figure~\ref{fig:entropy} and Figure~\ref{fig:seqlen}, during RLVR training, as the training steps increase, the model’s generated CoT sequences grow progressively longer. 
In the early stage, although the sequence length steadily increases, the model rapidly acquires stable short reasoning patterns, often accompanied by certain forms of spurious reasoning, resulting in a decrease in entropy. 
As training continues and CoT sequences further lengthen, the model begins to develop more adaptive reasoning due to exposure to perturbed data. 
This transition increases the model’s predictive uncertainty, leading to the subsequent rise in entropy.

\begin{figure}[b]
    \centering
    \includegraphics[width=\linewidth]{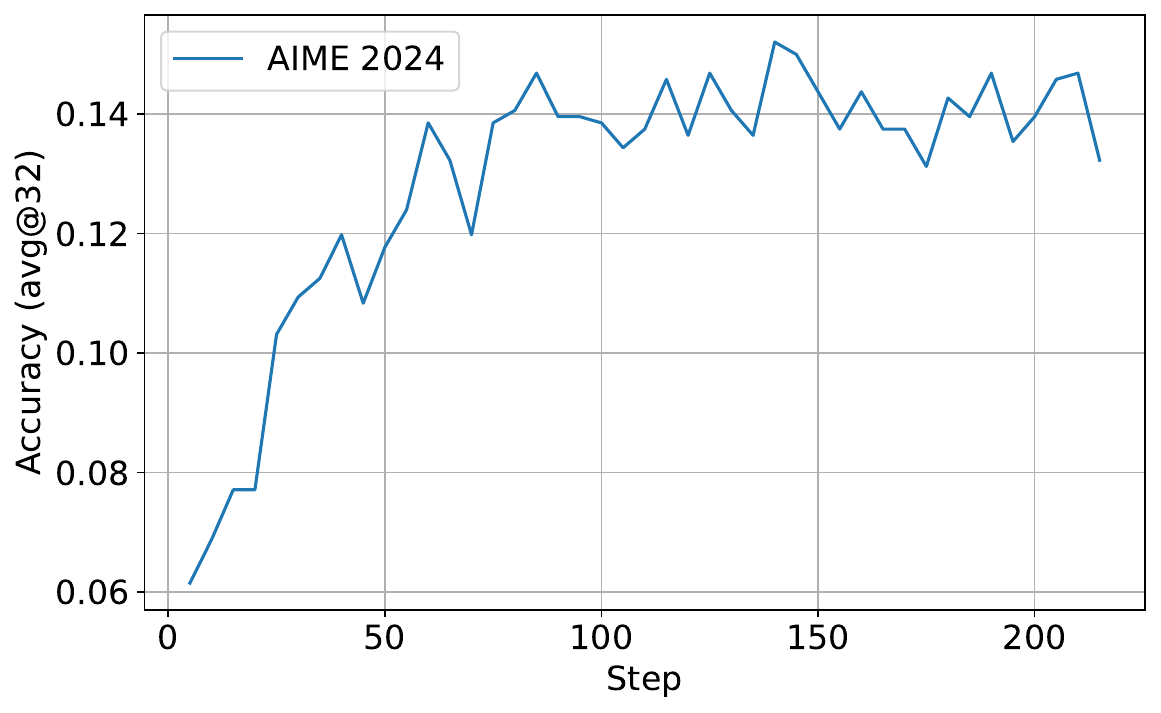}
    \caption{AIME 2024 scores on \nickname{} during training.}
    \label{fig:aime}
\end{figure}

In parallel, Figure~\ref{fig:aime} shows that the AIME 2024 scores exhibits a rapid improvement during the initial training stage and then plateaus with minor fluctuations in later steps. 
This trajectory aligns with entropy dynamics: the model first stabilizes around short reasoning patterns that yield quick performance gains, and later explores a broader reasoning space as CoT sequences grow, leading to small oscillations around a higher performance level.

% \newpage

\begin{table*}[htbp]
    \centering
    \resizebox{0.7\linewidth}{!}{
    \begin{tabular}{c c c c}
    \toprule
    Sanity Check  & Generation Times (s per 1K samples) & Retention rate (\%) \\
    \midrule
    Variable Alignment  & 2.90 & 67.70 \\
    Executable Code & 2.80 & 70.70 \\
    Existence of Valid Query & 612 & 95.17 \\
    \bottomrule
    \end{tabular}}
    \caption{Generation cost of component in \textit{sanity check}.}
    \label{tab:sanity_check}
\end{table*}

\begin{table*}[htbp]
\centering
\resizebox{\linewidth}{!}{
\begin{tabular}{c c c c c c c c c c c c}
\toprule
\multirow{2}{*}{Method} & \multirow{2}{*}{\# Samples} & \multicolumn{5}{c}{In-Domain} & \multicolumn{4}{c}{Out-of-Domain} & \multirow{2}{*}{AVG} \\
\cmidrule(lr){3-7} \cmidrule(lr){8-11} & & GSM8K & ORCA-{\nickname} & main & p1 & p2 & MATH & College & Theorem & AIME & \\
\midrule
Initial SFT & $0K$ & 81.43 & 77.12 & 74.26 & 64.48 & 52.44 & 42.84 & 28.55 & 14.88 & \ \ 4.27 & 48.92\\
\hdashline
\multirow{2}{*}{Qwen2.5-7B} & $9K$ &{87.34} & {83.72} & {81.02} & {73.84} & {62.08} & {64.36} & {35.42} & {25.13} & {13.33} & {58.47} \\
 & $9K\times 4$ &{89.31} & {85.36} & {88.50} & {80.72} & {70.76} & {70.70} & {41.34} & {30.13} & {14.90} & {63.52} \\
 \hdashline
\multirow{2}{*}{Qwen2.5-72B} & $9K$ & {91.81} & {86.08} & {89.72} & {82.74} & {73.24} & {75.90} & {48.62} & {37.13} & {14.27} & {66.61} \\
& $9K\times4$ & {92.72} & {87.16} & {90.32} & {83.50} & {73.80} & {76.20} & {49.96} & {37.50} & {16.46} & {67.51} \\

\bottomrule
\end{tabular}
}
\caption{
Performance comparison when employing different models for data synthesis.
}
\label{tab:data_syn}
\end{table*}

\begin{table*}[!htbp]
\centering
\resizebox{\linewidth}{!}{
\begin{tabular}{c c c c c c c c c c c}
\toprule
\multirow{2}{*}{Method} & \multicolumn{3}{c}{Robustness} & \multicolumn{5}{c}{Generalization} & \multirow{2}{*}{AVG} & \multirow{2}{*}{\begin{tabular}{c}CoT\\Length\end{tabular}}\\
\cmidrule(lr){2-4} \cmidrule(lr){5-9} & seen & unseen & GSM-SYM & GSM8K & MATH & College & Theorem & AIME & \\
\midrule
Qwen3-4B-Instruct-2507 & {60.99} & {56.62} & {85.50} & {92.27} & {91.12} & {50.85} & {47.50} & {46.46} & 66.41 & 1672 \\
Qwen3-4B-Thinking-2507 & {66.32} & {63.42} & {91.19} & {93.33} & \textbf{95.38} & \textbf{53.97} & \textbf{58.13} & \textbf{75.83} & \textbf{74.70} & 4784 \\
Standard-RLVR-Instruct & 62.86 & 57.72 & 90.47 & 93.78 & 91.52 & 52.09 & 48.25 & 52.50 & 68.65 & 2355\\
{\nickname}-Instruct & \textbf{68.42} & \textbf{65.17} & \textbf{91.88} & \textbf{94.69} & {90.83} & {53.30} & {55.00} & {58.83} & {72.27} & 3435 \\
\bottomrule
\end{tabular}
}
\caption{
Comparison with the Instruct model and Thinking model.
}
\label{tab:instruct_result}
\end{table*}

% \begin{table}[h]
% \centering
% \caption{
% Comparison with the target Instruct model.
% }
% \resizebox{\columnwidth}{!}{
% \begin{tabular}{c c c c c c c c c c}
% \toprule
% \multirow{2}{*}{Method} & \multicolumn{5}{c}{In-Domain} & \multicolumn{3}{c}{Out-of-Domain} & \multirow{2}{*}{AVG} \\
% \cmidrule(lr){2-6} \cmidrule(lr){7-9} & GSM8K & ORCA-{\nickname} & main & p1 & p2 & MATH & College & Theorem & \\
% \midrule
% Instruct & {92.57} & {81.44} & {89.60} & {81.86} & {70.68} & {79.86} & {51.70} & {44.25} & 73.99 \\
% {\nickname} & {91.81} & {86.08} & {89.72} & {82.74} & {73.24} & {75.90} & {48.62} & {37.13} & 73.16 \\
% {\nickname}-Instruct & {92.49} & {87.52} & {90.90} & {85.12} & {75.84} & {80.00} & {52.06} & {43.38} & \textbf{75.91} \\
% \bottomrule
% \end{tabular}
% }
% \label{tab:instruct_result}
% \vspace{-15pt}
% \end{table}

\clearpage 
\onecolumn
\twocolumn

\begin{promptbox*}{Generate Template and Code}
Task Description:

You are given a natural language query and its chain-of-thought response. 

Your task is to: 
Generate a Query Template by abstracting specific values into variables.

Generate Python Code that executes the logic described in the COT response using the abstracted variables.

\vspace{\baselineskip}
Input Format:

Query: Original query with specific values

Response: Chain-of-thought reasoning that leads to the answer

\vspace{\baselineskip}
Output Requirements:

Query Template:

Replace only concrete values in the query with angle-bracketed placeholders like \textless variable\_name\textgreater.
Do not replace names or general nouns (e.g., do not change ``Jungkook'' to \textless person\_name\textgreater).
Preserve the original wording and structure of the query as much as possible.

Python Code:

Begin by defining variables that correspond to the placeholders in the template.
Translate the logic in the response into executable Python code.
The code should end with a print() statement that prints only the final result.
Do not include comments with explanations or reasoning.
Use the same variable names as in the template for consistency.

\vspace{\baselineskip}
=== START EXAMPLE ===

\vspace{\baselineskip}
\{example\}

\vspace{\baselineskip}
=== END EXAMPLE ===

\vspace{\baselineskip}
\#\#\# Query:

\{query\}

\vspace{\baselineskip}
\#\#\# Response:

\{response\}

\end{promptbox*}

\clearpage

\begin{promptbox*}{Existence of Valid Query}
Task Description:

Your task is to generate a Chain-of-Thought (CoT) explanation that answers the user's question by reasoning through the logic implied in a provided Python script. Use the script to inform your explanation, but do not output or reproduce any code.

\vspace{\baselineskip}
Input Format:

Query: A question involving specific values or conditions.

Python Code: A script that solves the query or provides a key computational procedure.

\vspace{\baselineskip}
Output Requirements:

Start by interpreting the question clearly.
Reason through the problem step by step, using the Python code as a guide to inform your logic.
Refer to relevant steps in the code as part of your reasoning.
Do not output or reference the code in any form.
Explicitly state the final answer after the final step within \textbackslash boxed\{\}.

\vspace{\baselineskip}
\#\#\# Query:

\{query\}

\vspace{\baselineskip}
\#\#\# Python Code:

\{code\}

\end{promptbox*}

\begin{promptbox*}{Code Generation}
Please write a Python code to solve the following problem. Just give me the code, no explanation, no comments, no input statements. The code should be runnable and print the answer in the end.

\vspace{\baselineskip}
\#\#\# Query:

\{query\}

\vspace{\baselineskip}
\#\#\# Python Code:

\end{promptbox*}

\begin{promptbox*}{Code Execution}
Please help me run the following Python code and return its output result instead of the code itself:

\{code\}
\end{promptbox*}

\end{document}